\documentclass[10pt,twocolumn,letterpaper]{article}

\usepackage{cvpr}              %

\definecolor{cvprblue}{rgb}{0.21,0.49,0.74}
\usepackage[pagebackref,breaklinks,colorlinks,allcolors=cvprblue]{hyperref}
\usepackage{multirow}
\usepackage{amsmath}
\usepackage{wrapfig,booktabs}
\newcommand{\conditionalLink}[2]{%
  \ifcsname r@#1\endcsname
    \Cref{#1}%
  \else
    {#2}%
  \fi
}

\def\paperID{13902} %
\def\confName{CVPR}
\def\confYear{2025}

\title{Accurate Differential Operators for Hybrid Neural Fields}

\author{Aditya Chetan,
\
Guandao Yang,
\
Zichen Wang,
\
Steve Marschner,
\
Bharath Hariharan\\
Cornell University\\
{\tt\small achetan@cs.cornell.edu, \{gy46,\ zw336\}@cornell.edu, \{srm, bharathh\}@cs.cornell.edu}
}

\begin{document}

\twocolumn[{
    \renewcommand\twocolumn[1][]{#1}
    \maketitle
    \vspace{-2em}
    \begin{center}
        \vspace{-12pt}
    \centering
    \includegraphics[width=0.85\linewidth]{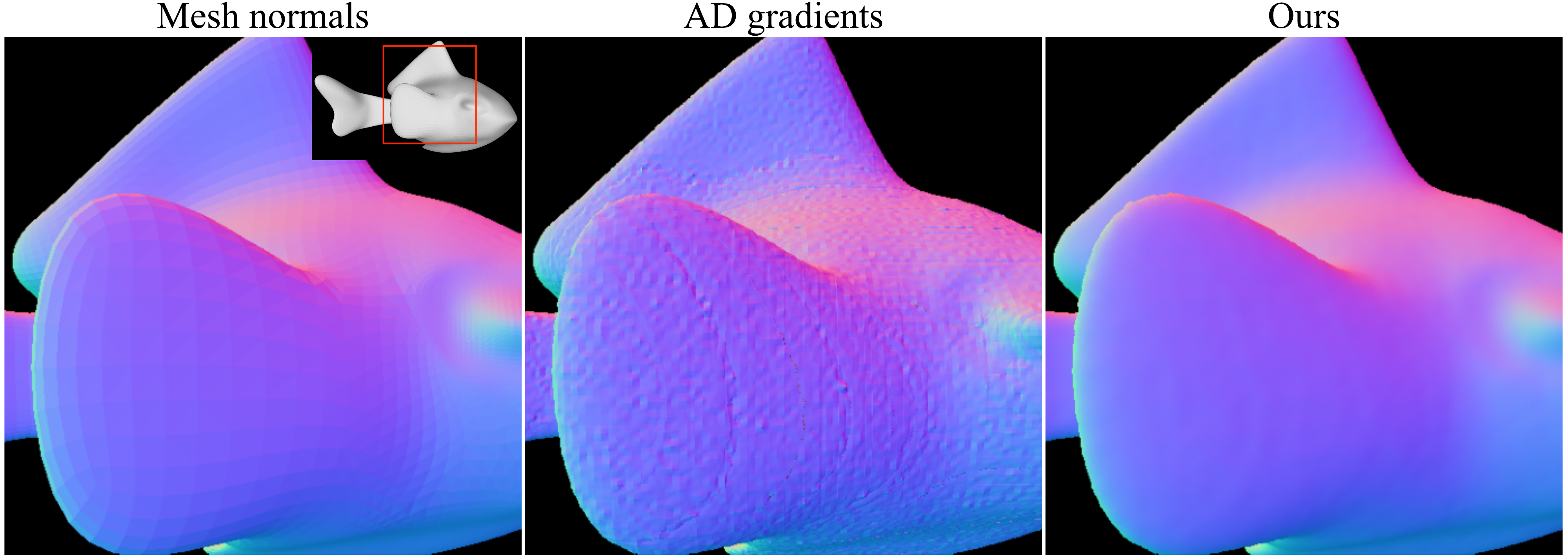}
    \captionof{figure}{\textbf{Noisy gradients in hybrid neural fields}. Normal images of \emph{Blub the fish}~\cite{blubthefish} (inset), using gradients queried from its hybrid neural SDF using automatic differentiation (AD) and our approach. Naively using AD gradients as surface normals leads to grainy artifacts, which our method alleviates.}\label{fig:independence}
    \vspace{10pt}

    \end{center}
    \vspace{-1.5em}
}]
\begin{abstract}
    Neural fields have become widely used in various fields, from shape representation to neural rendering, and for solving partial differential equations (PDEs). With the advent of hybrid neural field representations like Instant NGP that leverage small MLPs and explicit representations, these models train quickly and can fit large scenes. Yet in many applications like rendering and simulation, hybrid neural fields can cause noticeable and unreasonable artifacts. This is because they do not yield accurate spatial derivatives needed for these downstream applications. 
    In this work, we propose two ways to circumvent these challenges. Our first approach is a post hoc operator that uses local polynomial fitting to obtain more accurate derivatives from pre-trained hybrid neural fields. Additionally, we also propose a self-supervised fine-tuning approach that refines the hybrid neural field to yield accurate derivatives directly while preserving the initial signal. We show applications of our method to rendering, collision simulation, and solving PDEs. We observe that using our approach yields more accurate derivatives, reducing artifacts and leading to more accurate simulations in downstream applications. 
    \vspace{-2em}
\end{abstract}

\section{Introduction}

Neural fields are neural networks that take spatial coordinates as input and approximate spatial functions such as images~\cite{sitzmann2020siren}, signed distance fields~\cite{park2019deepsdf}, and radiance fields~\cite{mildenhall2020nerf}.
The advent of \emph{hybrid} neural fields, which modulate the neural network using features from a feature grid, has enabled much faster training~\cite{ingp,plenoxels,chen2022tensorf} and much better scaling to large-scale 3D structures with incredible detail, including entire cities~\cite{xiangli2021citynerf,tancik2022block,peng2020convolutional,takikawa2021nglod}.
Hybrid neural fields are thus gaining popularity as a representation of choice in many applications.
However, while these hybrid neural fields can be trained to represent large, complex spatial signals with high fidelity, we find that the \emph{derivatives} (computed with automatic differentiation or autodiff) of the trained field do not match the derivatives of the ground-truth signal;
\eg, compare the grainy normals obtained from a fully trained hybrid neural SDF to the much smoother normals from the mesh in~\Cref{fig:independence}.
Such artifacts in derivatives can cause significant artifacts in rendering \cite{takikawa2021nglod} or simulation pipelines \cite{chenwu2023insr-pde} which heavily rely on accurate derivatives.
Thus, for hybrid neural fields to fulfill their promise of a practical representation for spatial signals, we need to eliminate these errors in the derivatives.

\vspace{-0.5em}
Why are the derivatives of hybrid neural fields so noisy?
We observe that to enable the capture of complex geometry with high fidelity, hybrid fields are designed to have high-frequency components (\eg, spatial grids of a high resolution).
As such, they also have high-frequency noise.
This noise will be of fairly low magnitude in a well-trained hybrid field.
But even so, differentiation
will significantly amplify this high-frequency noise (\Cref{sec:polyfit}) resulting in the artifacts that we see.
We posit that we need a new differentiation operator that is robust to high-frequency noise.

In this paper, we propose a new approach to reduce noise in the derivatives of pre-trained hybrid neural fields.
Our approach takes inspiration from classical signal processing where derivatives are typically done on a smoothed version of the signal to avoid amplifying high-frequency noise.
Our key idea is to replace direct derivatives of the hybrid neural field with derivatives of a \emph{local} low-degree polynomial approximation of the field.
These low-degree polynomials can be fit in closed form and effectively remove high-frequency noise.
Importantly, this approach is general and can apply to any hybrid neural field independent of architecture.

While this approach yields accurate derivatives for off-the-shelf neural fields, it requires that downstream pipelines be changed to use our new derivative operator.
To avoid altering downstream pipelines, we propose an extension of this approach where we use the accurate derivatives from the low-degree local polynomial fit to regularize the neural field during training/finetuning.
Concretely, we add an auxiliary loss that penalizes the difference between the autodiff gradients of the neural field and the derivatives from the local polynomial approximation.
This yields a new hybrid neural field where autodiff itself yields accurate derivatives.

Our experimental results show that our new derivative operator yields more accurate derivatives than autodiff, reducing errors in gradients by \textbf{4$\times$}. It also outperforms other alternative derivative operators, such as finite difference stencils, reducing errors in curvature by \textbf{4$\times$}.
We also show that using our operator to regularize neural field finetuning improves derivative accuracy, outperforming other regularization strategies that encourage smoothness like eikonal regularization~\cite{atzmon2019sal, gropp2020implicit,li2023neuralangelo},  showing that existing approaches are not well-suited for hybrid fields.
Lastly, we demonstrate that our approaches substantially reduce artifacts in downstream rendering and simulation applications. 
Thus our proposed methods open the door for using hybrid neural fields in a large set of downstream applications.
\vspace{-1.2em}
\paragraph{Contributions.} Our overall contributions can be summarized as follows: (1) We identify the issue of inaccurate derivatives in a given pre-trained hybrid neural field and point out its relationship to high-frequency noise. (2) We propose a local polynomial-fitting operator to improve the accuracy of neural field derivatives. (3) We also propose a fine-tuning approach to improve the quality of autodiff derivatives of hybrid neural fields. 
We provide an implementation of our operators at: {\small\url{https://justachetan.github.io/hnf-derivatives/}}

\section{Related Work}

\paragraph{Neural Fields.}
Neural fields are neural networks approximating spatial fields given coordinates as input~\cite{yang2021geometry,xie2022neural}.
They have been used to represent megapixel images~\cite{martel2021acorn}, 3D shapes in implicit fields~\cite{park2019deepsdf,mescheder2019occupancy,chen2019learning} and radiance fields ~\cite{mildenhall2020nerf,barron2021mip,Verbin2021RefNeRFSV}.
Our work is applicable to hybrid neural fields in all these applications, although our primary evaluation is on SDFs.
Typical neural field architectures are multi-layer perceptrons~\cite{sitzmann2020siren,tancik2020fourier,mildenhall2020nerf}, but these can be slow to train and may not scale to large scenes with fine-grained details.
As such, more current approaches use 
hybrid representations that modulate an MLP with spatial features stored on a grid~\cite{ingp,takikawa2021nglod,yu2021plenoctrees,fridovich2022plenoxels,chen2022tensorf}.
These hybrid techniques scale well ~\cite{peng2020convolutional,xiangli2022bungeenerf,tancik2022block}, but we show that they yield noisy derivatives: the key issue we strive to address here.
Accurate derivatives are particularly important when neural fields are used for applications such as rendering~\cite{takikawa2021nglod,wang2021neus,iron-2022} and simulation~\cite{sitzmann2020siren,chenwu2023insr-pde,li2023pac,chen2023crom}. 
Recently, similar to our approach, Li~\etal~\cite{li2023neuralangelo} used a finite differences-based regularizer for training hybrid neural fields for surface reconstruction. However, their motivation is to address the training dynamics of hybrid fields, instead of removing their high-frequency noise components. Additionally, past approaches for reconstructing surfaces from point clouds~\cite{ben2022digs,OnSurfacePriors,PredictiveContextPriors} also include regularization terms that could potentially lead to more accurate derivatives. However, they are specifically designed for non-hybrid neural fields like SIREN~\cite{sitzmann2020siren} with architecture-specific initialization and higher-order loss functions. In contrast, our approach targets hybrid neural fields like Instant NGP~\cite{ingp}. It is also non-trivial to apply these approaches to improve the spatial derivatives of \textit{pre-trained} hybrid neural fields. 

\vspace{-1.2em}
\paragraph{Numerical Derivatives of Noisy Signals.} Estimating derivatives of noisy signals is a classical problem in numerical differentiation. Previous works \cite{ejde-v21-knowles2014} propose different approaches to regularize the noise, such as total variation minimization, Tikhonov regularization, convolution smoothing, etc., depending on the type of noise model applied. However, these approaches are typically limited to computing derivatives on a uniform grid. In contrast, our operators can query the derivative at any arbitrary point in a continuous space; a desirable flexibility in downstream applications such as computing differential operators on 3D shapes. 
Furthermore, these methods are designed to compute derivatives on 1D~\cite{ejde-v21-knowles2014} or 2D grids~\cite{twodTV,VillaPetraGhattas16}, and scaling them to higher dimensions like 3D is a non-trivial extension. 
Our work is also closely related to past works in differentiable rasterization~\cite{fischer2023plateau,deliot2024Transform} that estimate the derivatives of noisy signals using Monte Carlo estimation often by smoothing them first with a Gaussian kernel. If the signal is non-differentiable, they convolve the signal with a derivative-of-Gaussian filter or differentiate a locally-fitted differentiable surrogate signal~\cite{fischer2024zerograds}.

\vspace{-1em}
\paragraph{Polynomial fitting for Shape Analysis.} Polynomial-fitting approaches like Moving Least Squares (MLS)~\cite{2004shortmlsintro,2008ChengMLSSurvey,2000LevinMLS} have a rich history in 3D shape analysis. They have applications in tasks like surface reconstruction from point clouds \cite{alexa2003tvcg}, animating elastoplastic materials~\cite{muller2004mlsanimation}, and learning implicit functions from scattered data~\cite{2004ShenImplicitMLS,2003OhtakePartitionImplicitsMLS}. In this paper, we apply polynomial fitting to a novel setting of hybrid neural fields to solve the important issue of obtaining accurate differential operators. Typically, in past works, given scattered data (point clouds with associated scalar values) as input, approaches like MLS compute fitting planes (or higher-order polynomials) to local subsets of surface points. In essence, the planes/polynomials serve as an \textit{interpolant} for the given data (point clouds). In our setting, the hybrid neural field \textit{already exists} as an interpolant. But, as we observe in \Cref{fig:independence}, the neural field interpolant does not yield accurate derivatives, and our approach attempts to alleviate this problem.

\section{Method}
\label{sec:method}

\begin{figure*}
    \begin{subfigure}{0.52\textwidth}
        \centering
        \includegraphics[width=\textwidth]{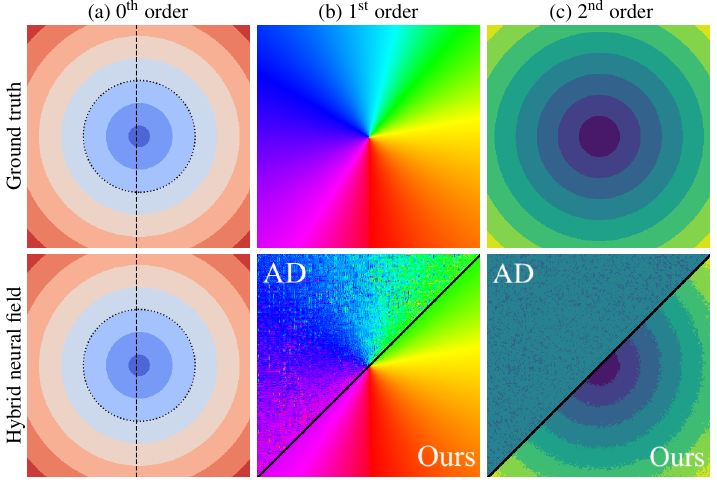}
        \caption{}
        \label{fig:motivation}
    \end{subfigure}%
    \hspace{1mm}
    \begin{subfigure}{0.45\textwidth}
        \centering
        \includegraphics[width=\textwidth]{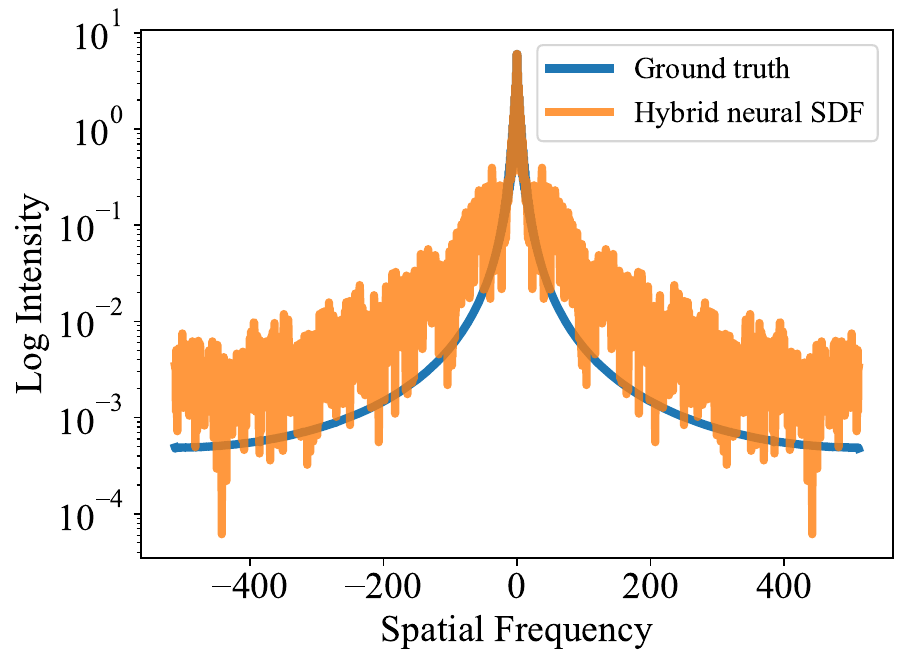}
        \caption{}
        \label{fig:fft}
    \end{subfigure}
    \vspace{-1em}
    \caption{\textbf{(a) Inaccurate differential operators of hybrid neural fields.} Hybrid neural SDF of a circle in 2D. As shown by the comparison with the ground truth, the $0^{\mathrm{th}}$ order signal accurately captures the SDF. But the $1^{\mathrm{st}}$ and $2^{\mathrm{nd}}$-order signals, here shown as the gradient and the radius of curvature (inverse of the Laplacian) are quite noisy. \textbf{(b) Fourier spectrum of a hybrid neural SDF.} Computed over a 1D slice (dashed line in (a)) of the SDF of a 2D circle. Note the noisy high-frequency components that are captured by the hybrid neural field.}
    \vspace{-1.5em}
\end{figure*}

We assume that we have a \textit{pre-trained} neural field, $F_\theta$.
To concretize the problem, we focus on hybrid neural fields representing 3D shapes as signed distance fields (although our final approach is more general and applicable to other modalities too, see \conditionalLink{sec:addlresultsimages}{Appendix B.2}).
By \textit{hybrid} fields~\cite{ingp} we refer to neural fields that have a spatial grid of feature vectors in addition to an MLP. 
The field value at any point is obtained by feeding to the MLP the point location as well as a feature vector obtained by interpolating into the grid.
We begin by analyzing why hybrid neural fields yield noisy derivatives and then motivate our approach.

\subsection{Noisy Derivatives in Hybrid Neural Fields}
\label{sec:polyfit}

Why are the derivatives of hybrid neural fields incorrect?
We observe that much of the capacity of hybrid neural fields lies in the high-resolution spatial grid of feature vectors.
This spatial grid is essential for the neural field to capture fine-grained localized details.
Consequently, this spatial grid also determines the high-frequency components of the fitted signal. 
Unfortunately, this abundance of capacity for high-frequency components means that there are likely many solutions with different high-frequency components that fit the training data well.
This in turn can result in noise in the high-frequency components.
We observe this noise in practice. \Cref{fig:fft} compares the spectrum of the ground truth and learned signed distance function (SDF) for a circle in 2D. Note how the learned SDF has higher amplitudes in the high-frequency components.

This high-frequency noise is the source of artifacts in the derivatives.
This is because derivative computation accentuates high-frequency noise, scaling it up proportional to the frequency, as illustrated by a sinusoidal signal with frequency $\nu$: $
\frac{d \sin(2\pi\nu x)}{dx} = 2\pi\nu cos(2\pi\nu x) 
$.
Thus, even when the high-frequency noise has a very low magnitude, the corresponding noise in the derivative has a much higher magnitude.
\Cref{fig:motivation} shows this issue in practice: the same SDF of a 2D circle that we learned earlier provides an extremely noisy gradient when we use automatic differentiation.
\vspace{-1em}
\paragraph{Derivatives and smoothing.}
This notion of high-frequency noise magnifying errors in derivative computation is well-known in signal processing, and the solution is to use smoothing to remove the high-frequency components.
\begin{figure}
    \centering
    \includegraphics[width=0.48\textwidth]{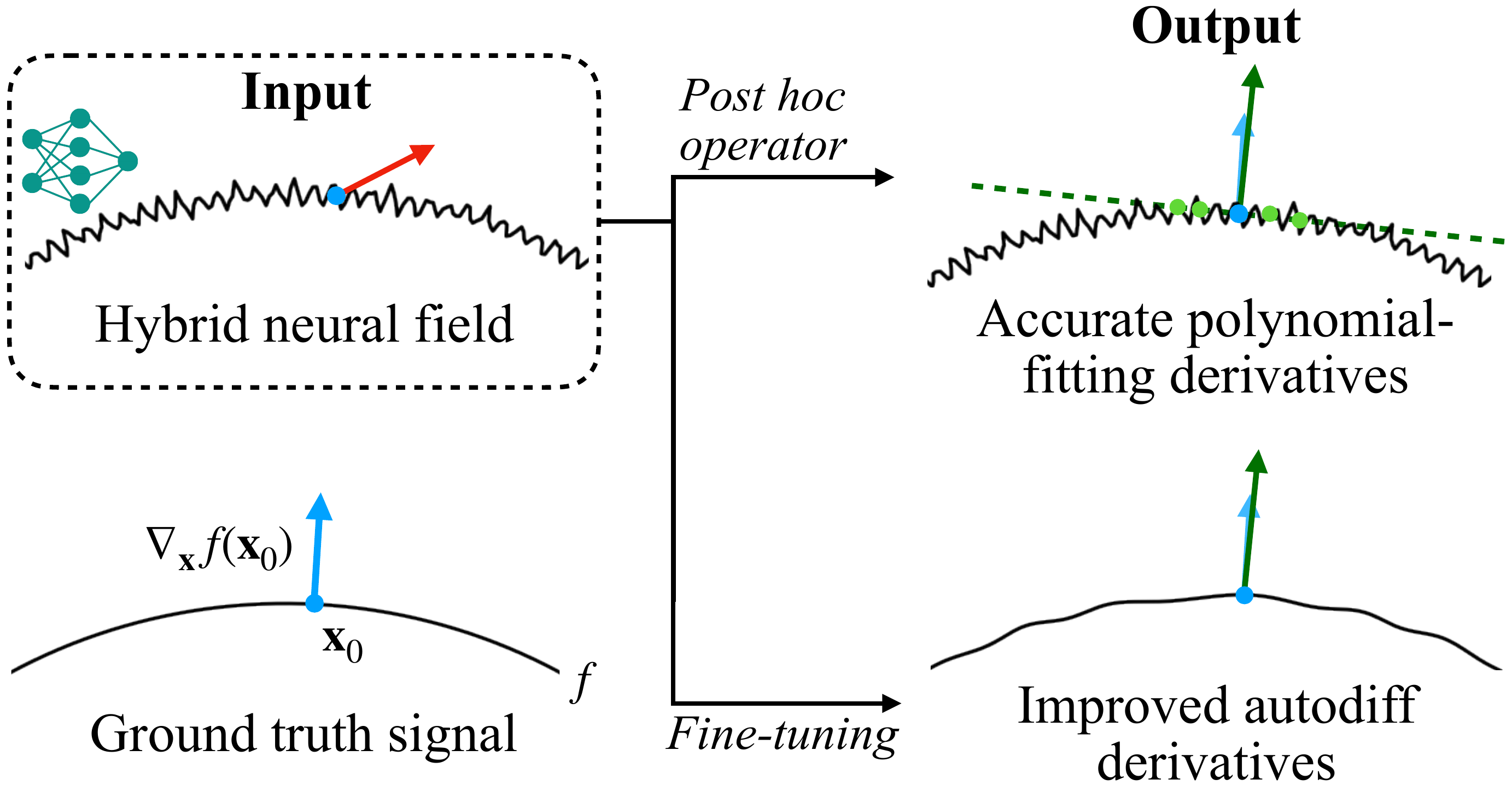}
    \caption{\textbf{Problem setup.} Given a pre-trained hybrid neural field with noisy autodiff derivatives, we propose two approaches for accurate derivatives. Our polynomial-fitting operator can be applied in a post hoc manner while our fine-tuning approach directly improves autodiff derivatives of the field.}
    \label{fig:setup}
    \vspace{-1.5em}
\end{figure}

The degree of smoothing can be controlled and corresponds to the scale of the derivative.
How this smoothing is done depends on how the signal is represented.
For images represented as a 2D grid of pixel values, smoothing can be done by convolving with an averaging filter, and derivatives are typically only computed after smoothing.
When 3D shapes are represented as meshes, the mesh automatically represents a smooth version of the signal: each face is effectively a local linear approximation of the surface.
Derivatives can then be computed using the face normal.

Unfortunately, no analogous notion of smoothed derivatives exists for arbitrary hybrid neural fields.
To address this gap, we propose a new approach that computes derivatives on a local low-degree polynomial approximation of the neural field.
We describe this approach in detail below (see \Cref{fig:setup} for an overview).

\subsection{Local polynomial-fitting operators}
\label{sec:localpolyfit}

Given a hybrid neural field, $F_\Theta: \mathbb{R}^m \rightarrow \mathbb{R}^n$, and a query point $\mathbf{q} \in \mathbb{R}^m$, we want to compute accurate first-order derivatives of $F_\Theta$ at $\mathbf{q}$. 
For simplicity, we choose $n=1$.

First, we sample points $\mathbf{x}_i, i = 1, \ldots, k$ from a local neighborhood $N(\mathbf{q})$ of the point $\mathbf{q}$. 
We query the neural field to obtain corresponding field values $y_i = F_\Theta(\mathbf{x}_i)\ \forall \mathbf{x}_i \in N(\mathbf{q})$. We then use these values to fit a local linear approximation $y \approx \hat{F}_{\Theta}(\mathbf{x}; \mathbf{q}) = \mathbf{g}^T \mathbf{x} + b$ using simple least squares:
\vspace{-1em}
\begin{align}
\hat{\mathbf{g}}, \hat{b} = \arg \min_{\mathbf{g}, b} \sum_{i=1}^k (\mathbf{g}^T \mathbf{x}_i + b - y_i)^2
\end{align}
Our estimate of the derivative is then $\hat{\nabla}_{\mathbf{x}} F_\Theta (\mathbf{q}) = \nabla_{\mathbf{x}} \hat{F}_{\Theta}(\mathbf{x}, \mathbf{q}) = \hat{\mathbf{g}}$.
We can extend the same approach to the case of vector fields ($n > 1$), where $\mathbf{g}$ is replaced by an $m \times n$ estimate of the Jacobian, $\mathbf{J}$.
Observe that this optimization problem is an unconstrained convex quadratic program that can be solved in closed form.

\vspace{-1em}
\paragraph{Local neighborhood selection.}

Different sampling schemes can be considered in order to select a local neighborhood around the query point $\mathbf{q}$. However, in our experiments, we found that sampling from a Gaussian distribution centered at $\mathbf{q}$, $\mathcal{N}(\mathbf{q}, \sigma)$ worked best for us. The standard deviation, $\sigma$ controls the amount of smoothing that we do at a particular point. The number of neighbors sampled, $k$ is another hyperparameter of our method and controls the variance of the operator that we compute. We discuss how we select these hyperparameters in detail in our experiments (\Cref{sec:experiments}).

\vspace{-1em}
\paragraph{Hessian \& Laplacian.}

To compute second-order differential operators like the Hessian or Laplacian,  we fit a quadratic approximation in the neighborhood of $\mathbf{q}$ as opposed to a linear one.
Specifically, for scalar fields, we minimize: $\sum_{i=1}^k (\mathbf{x}^T_i \mathbf{H} \mathbf{x}_i + \mathbf{p}^T\mathbf{x}_i + q - y_i)^2$.
Ideally, since the Hessian is symmetric and $\mathbf{H}$ is our estimate for the Hessian, we want $\mathbf{H}$ to be symmetric. We therefore parameterize $\mathbf{H}$ by its lower triangle.
As before, this quadratic program can be solved in closed form.
Once we obtain $\mathbf{H}$, we can also obtain the Laplacian ($\Delta F_\Theta$) as the trace of $\mathbf{H}$.

Given any pre-trained neural field with similar high-frequency noise, our operators can be applied to it in a post hoc manner to obtain accurate differential operators from the field. However, they do not alter the weights of the neural field, essentially acting as ``test-time" operators. 

\vspace{-1em}
\paragraph{Comparison to alternatives.} Our approach computes the derivative by sampling points locally and fitting a local polynomial approximation.
However, one might consider other alternatives:
\begin{enumerate}
\item Instead of autodiff, which yields the instantaneous derivative, we can compute derivatives using finite differences. However, this amounts to sub-sampling the signal without smoothing, which will cause aliasing and thus, inaccuracy in derivatives, as we demonstrate in our experiments (see \Cref{sec:experiments}).
\item A mesh also computes a local polynomial approximation, so we could convert the neural field to a mesh using Marching Cubes. 
However, computing a mesh is a global operation, as opposed to our polynomial fit which can be solved in closed form independently for every query point. 
As such, extracting a mesh is much more expensive, especially for applications like physical simulation where each simulation step may require gradient queries from an evolving signal (see \conditionalLink{sec:mccmpr}{Appendix C}). %
\end{enumerate}

\vspace{-1.5em}
\subsection{Fine-tuning pre-trained hybrid neural fields} 
\label{sec:ft}
The post hoc operator we describe above can be used to effectively query accurate differential operators from a given hybrid neural field.
However, to use it, every downstream application must be altered to allow for our new operator.
Unfortunately, for many applications, autodiff remains the prevalent way to obtain gradients from neural networks. 
Hence, we propose a method to update the hybrid neural field directly so that autodiff yields accurate gradients.

Concretely, given a pre-trained neural field, we propose to fine-tune it to improve the accuracy of the differential operators obtained using autodiff. Let us denote the pre-trained neural field and the fine-tuned neural field by $M$ and $F_\Theta$ respectively. $F_\Theta$ is initialized with the weights of $M$. We fine-tune $F_\Theta$ using the following loss function:
\vspace{-0.5em}
{\small
\begin{align}
\begin{split}
    \mathcal{L}_{ft}(\mathbf{x}_0; \Theta) = & \underbrace{|F_\Theta(\mathbf{x}_0) -  M(\mathbf{x}_0) |^2}_{\mathcal{L}_{\mathrm{con}}}\\
    & + \underbrace{||\nabla_{\mathbf{x}}F_\Theta(\mathbf{x}_0) - \hat{\nabla}_{\mathbf{x}}M(\mathbf{x_0})||_2^2}_{\mathcal{L}_{\mathrm{grad}}}\label{eq:ftloss}
\end{split}
\end{align}
}%
Here, $\mathcal{L}_{\mathrm{con}}$ denotes the consistency loss which ensures that the output of $F_\Theta$ matches the pre-trained neural field, $M$. $\mathcal{L}_{\mathrm{grad}}$ denotes the gradient loss that tries to align the autodiff gradient of $F_\Theta$ with accurate gradient estimates obtained by applying the operator $\hat{\nabla}_{\mathbf{x}}$ on $M$. In our experiments, we use our polynomial-fitting gradient operator to obtain $\hat{\nabla}_{\mathbf{x}}M$. 

Our approach resembles the Sobolev training approach proposed in Yuan~\etal~\cite{yuan2022sobolev} with the distinction that they assume access to the ground-truth derivatives of the input signal, whereas we only assume access to noisy gradients of the pre-trained neural field.
Note that this fine-tuning process is orthogonal to any kind of smoothed gradient operator. Our polynomial-fitting gradient for $\hat{\nabla}_{\mathbf{x}}$ is just one of the ways we can perform this fine-tuning. We can similarly use other approaches to compute accurate gradient estimates.
In fact, in our experiments, we find that even less accurate estimates, like those obtained from finite differences, can suffice to regularize the fine-tuning effectively.  

We can also use the loss function described in \cref{eq:ftloss} as an auxiliary regularizer when training a hybrid neural field from scratch. In this case, we train the model ($F_\Theta$) normally using MSE loss with the ground truth SDF initially for $s\ (> 0)$ steps. This \textit{warm-start} phase allows $F_\Theta$ to learn a good initial fit for the zeroth-order signal. Then we train $F_\Theta$ with the loss in \cref{eq:ftloss} for $n-s$ steps where $n$ is the total number of training steps. $M$ is the frozen weights of $F_\Theta$ at the end of $s$ steps. The choice of $s$ plays an important role in the accuracy of autodiff gradients of the resulting model (see \conditionalLink{app:latereg}{Appendix F} for a discussion).

\section{Experiments and Results}
\label{sec:experiments}

We first evaluate the accuracy of our proposed operator and then evaluate our fine-tuning approach.
For both sets of experiments, we use shapes from the FamousShape dataset \cite{ErlerEtAl:Points2Surf:ECCV:2020}. 
We pre-train a hybrid neural field to learn the SDF of each shape. We experimented with \textbf{three hybrid architectures}: \textbf{Instant NGP 
}\cite{ingp}, Instant NGP without a hash grid (\textbf{Dense Grid}), and \textbf{Tri-plane} \cite{Chan2021Triplane}.

\vspace{-1em}
\paragraph{Metrics.}
We evaluate the estimates of surface normals (first-order operator) and mean curvatures (second-order operator) by comparing them to surface normals and discrete mean curvatures obtained from the provided meshes of the shapes (which we regard as ground truth, see \conditionalLink{Appendix A}{sec:expdetails} for details). 
For surface normals, we compute the mean L2 error, mean angular error in degrees (Ang), and the percentage of points having angle error below $1^{\circ}$ (AA@1) and $2^{\circ}$ (AA@2). For mean curvature, we use the rectified relative error (RRE) used by past works for evaluating curvature estimation \cite{GuerreroEtAl:PCPNet:EG:2018,shabat2020deepfit}.
We report metrics averaged over all evaluated shapes (detailed results in \conditionalLink{sec:addlaccanalysis}{Appendix B.1}). For the detailed experimental setup, please refer to \conditionalLink{sec:expdetails}{Appendix A}. 

\begin{figure}
    \centering
    \includegraphics[width=\linewidth]{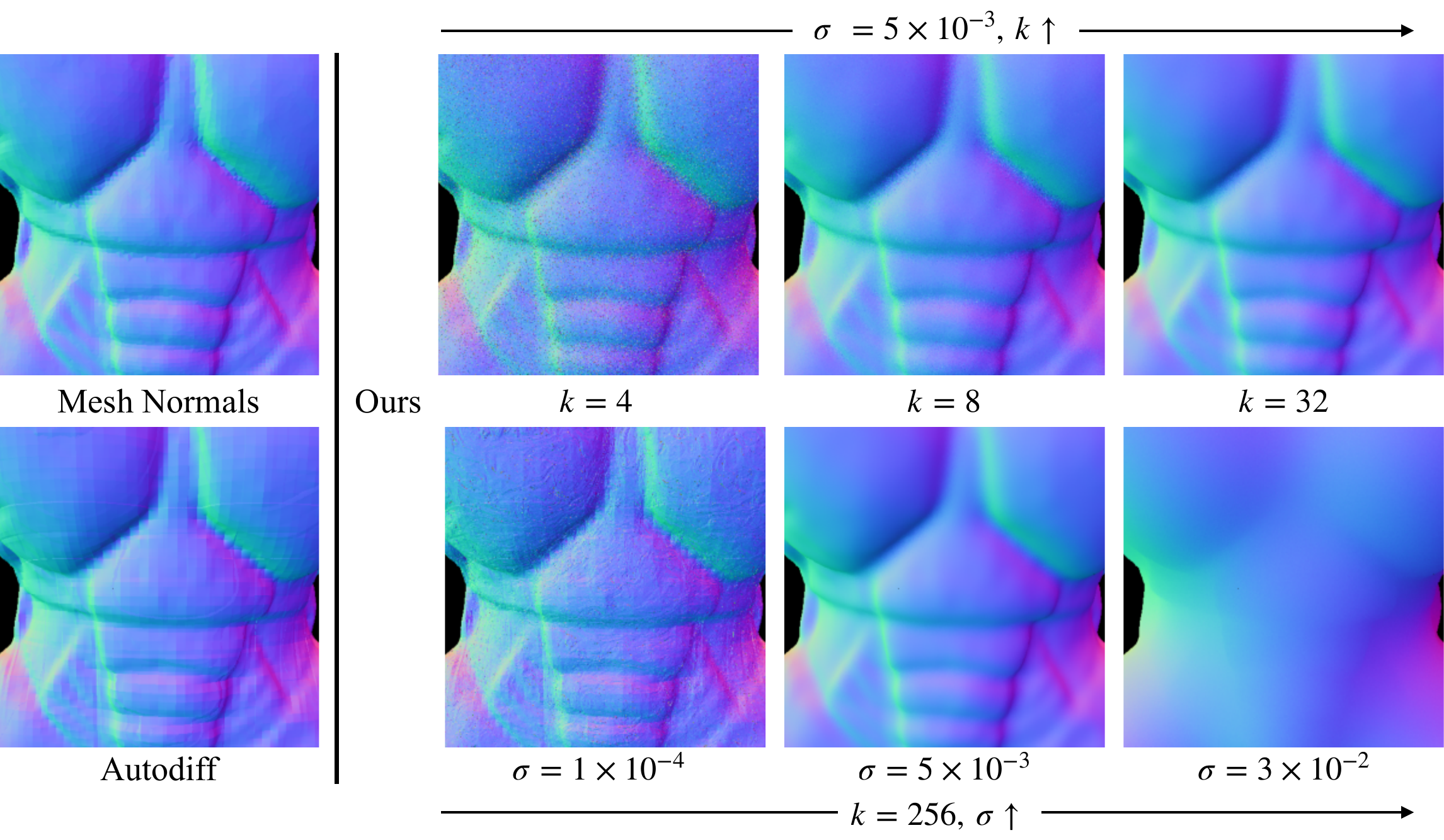}\\
    \includegraphics[width=\linewidth]{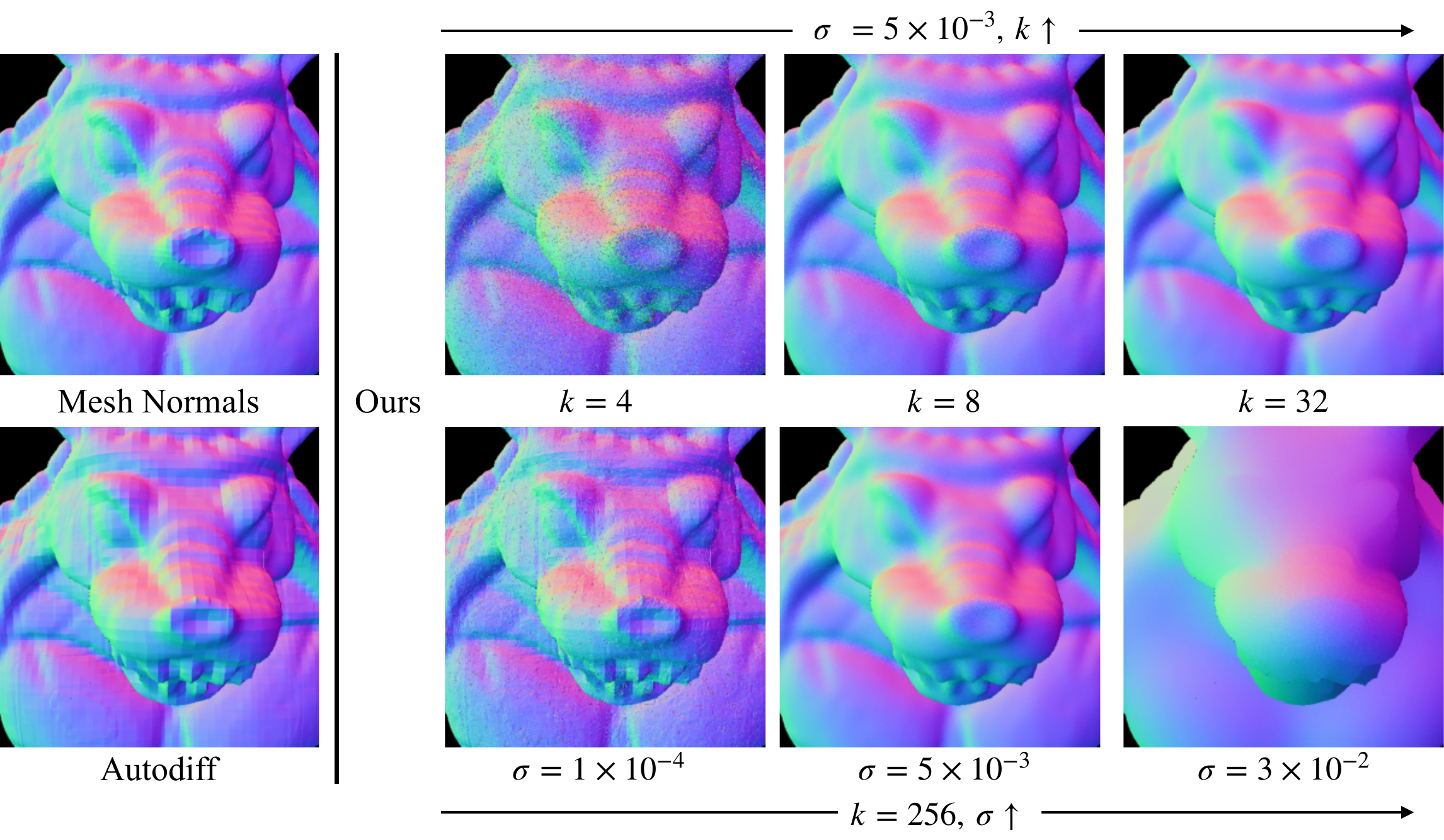}
    \vspace{-2em}
    \caption{\textbf{Effect of hyperparameters.} The performance of our polynomial-fitting operator is influenced by the selected hyperparameter values. We demonstrate this on the Armadillo highlighting this from two different viewpoints, the torso and the head. For a fixed $\sigma$, choosing a larger $k$ reduces the variance in our operator leading to smoother normals (see the first row for each viewpoint). For a fixed $k$, choosing a large $\sigma$ can lead to over-smoothing, whereas choosing a smaller $\sigma$ can lead to no smoothing at all (second row of each viewpoint). Best viewed by zooming in.}
    \label{fig:hparameffect}
    \vspace{-1.5em}
\end{figure}

\begin{figure*}[h!]
  \centering
  \begin{subfigure}[b]{0.45\linewidth}
    
    \includegraphics[width=\linewidth]{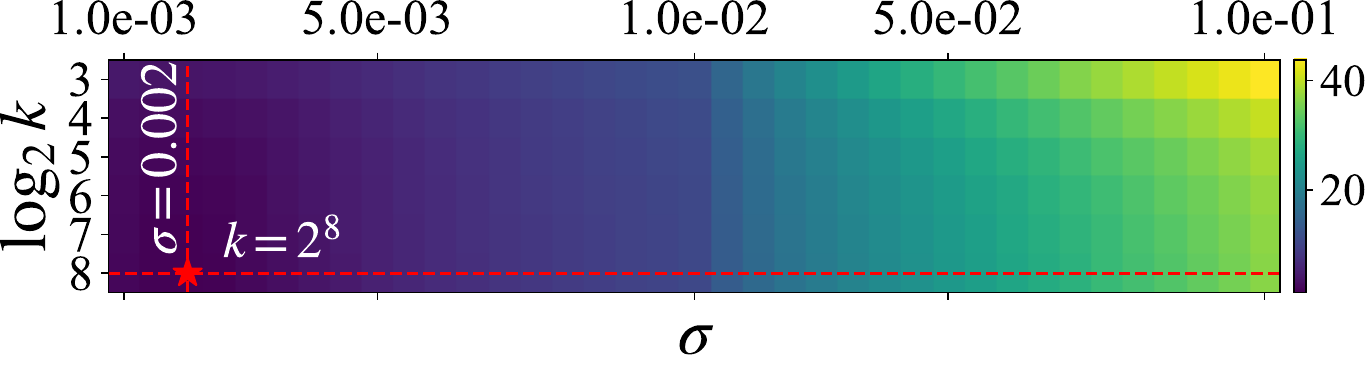}
    \includegraphics[width=\linewidth]{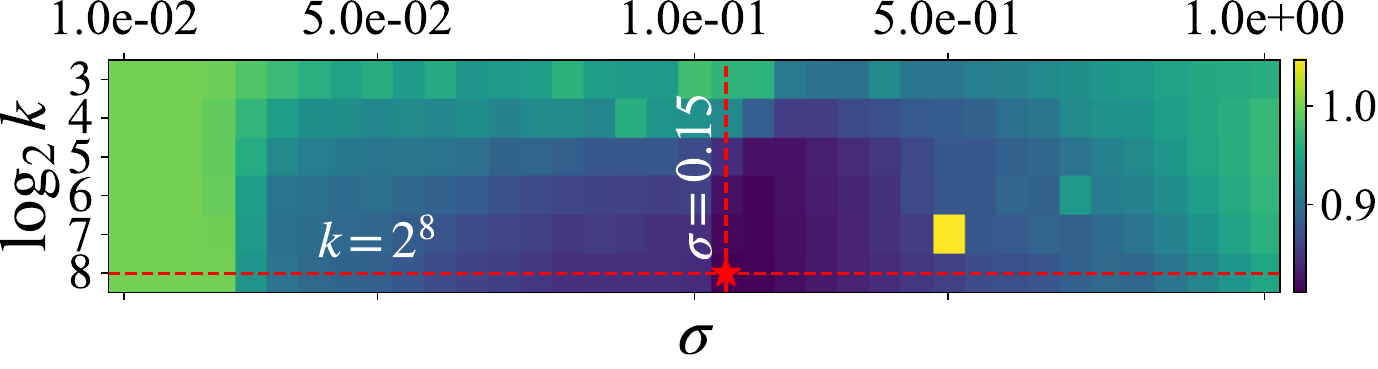}
    \caption{Armadillo}
  \end{subfigure}
  \begin{subfigure}[b]{0.45\linewidth}
    \includegraphics[width=\linewidth]{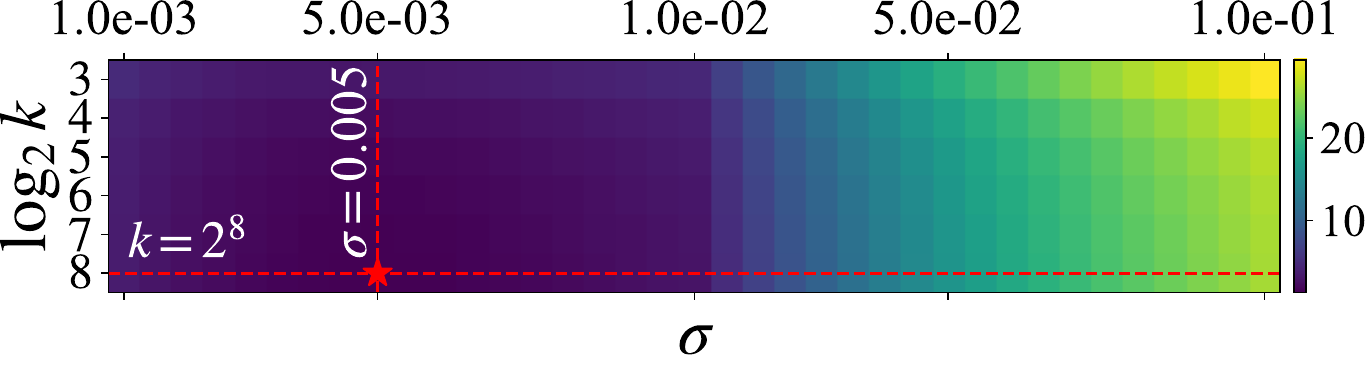}
    \includegraphics[width=\linewidth]{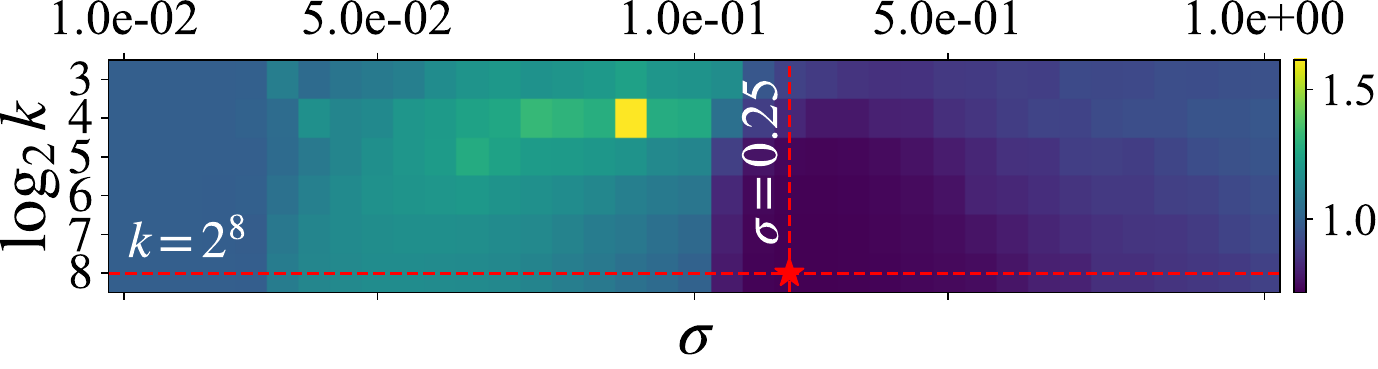}
    \caption{Stanford Bunny}
  \end{subfigure}  
  \vspace{-0.5em}
  \caption{\textbf{Hyperparameter Ablation}. Variation in angle error for normal/gradient (top) and the mean curvature error (bottom) for different settings. $k$ and $\sigma$ refer to the number of neighbors sampled and the size of the neighborhood respectively. $\star$ denotes the best settings.}
  \label{fig:hparamablation}
  \vspace{-1.5em}
\end{figure*}

\vspace{-1em}
\paragraph{Choosing $\sigma$ and $k$.} As discussed in \Cref{sec:localpolyfit}, our polynomial-fitting operators also require $\sigma$ and $k$ values as hyperparameters. 
The effect of these hyperparameter choices is shown qualitatively in \Cref{fig:hparameffect} and quantitatively in \Cref{fig:hparamablation} on the Armadillo and Bunny shapes.
Generally, we find that (a) higher $k$ (more neighbors) are always better as this minimizes variance, and (b) no single value of $\sigma$ works for both shapes, but derivative accuracy varies smoothly with $\sigma$.
Intuitively, $\sigma$ trades off between fidelity and robustness to noise. 
As such, it is dependent on the nature of the downstream application. 

For the purpose of our experiments, we always choose $k=256$. 
We choose $\sigma$ to have the best consistency with differential operators obtained from the mesh. Specifically,
\begin{itemize}
    \item For post hoc operators, we do a telescopic search for the best value of $\sigma$.
    \item For fine-tuning, we train an ensemble of models with different values of $\sigma$ and select the value that yields the best autodiff gradients after fine-tuning.
\end{itemize}

\subsection{Accuracy of operators.} 

We first evaluate our polynomial-fitting operator by comparing it with automatic differentiation, finite differences (FD) baseline, a stochastic finite differences operator~\cite{deliot2024Transform} (SFD) and a Monte Carlo estimate that aggregates information from samples in local neighborhoods~\cite{fischer2023plateau} (GAD). Specifically, GAD does Gaussian averaging of autodiff derivatives with importance sampling, mathematically equivalent to convolution with a derivate-of-Gaussian filter~\cite{fischer2023plateau}. Since SFD is a high-variance approach, we also compared against Monte Carlo averaging of SFD with 256 samples ($\text{SFD}_{\text{256}}$).
\Cref{tab:opacc} shows our results. We only compare FD and our approach for mean curvature, since our hybrid neural fields do not admit meaningful higher-order spatial gradients through autodiff (as they are piecewise linear) and Deliot~\etal~\cite{deliot2024Transform} do not discuss an SFD curvature operator. Our approach provides more accurate surface normals and mean curvature values from hybrid neural fields than the FD baseline. In particular, for Instant NGP \cite{ingp} our approach yields $\mathbf{4}\times$ reductions in the angular error for the surface normal, relative to the commonly used FD approach. Our approach performs comparable to GAD, showing that aggregating function values in local neighborhoods whether using polynomial fitting or Monte Carlo estimates can effectively address the high-frequency noise in hybrid neural fields. Our approach also yields higher accuracy when computing mean curvature relative to finite differences, leading to $\mathbf{4}\times$ reduction in error for Instant NGP~\cite{ingp}. 

\begin{table}[h!]
    \centering
    \renewcommand{\arraystretch}{1.2}
    \renewcommand{\tabcolsep}{1.2mm}
    \resizebox{\linewidth}{!}{
    \begin{tabular}{ccccccc}
        \toprule
        Model & Method & \multicolumn{4}{c}{Surface Normal} & Mean Curvature \\
        \cmidrule(lr){3-6} \cmidrule(lr){7-7}
        & & L2 $\downarrow$ & Ang $\downarrow$ & AA@1 $\uparrow$ & AA@2 $\uparrow$ & RRE $\downarrow$ \\
        \midrule
        \multirow{6}{*}{Instant NGP \cite{ingp}} & AD & 0.21 & 12.40 & 1.58 & 6.12 & - \\
        & FD & 0.07 & 4.20 & 26.86 & 55.22 & 3.67 \\
        & GAD & \textbf{0.05} & \underline{2.99} & \underline{38.35} & \underline{66.86} & - \\
        & SFD & 0.95 & 57.67 & 0.01 & 0.07 & - \\
        & $\text{SFD}_{\text{256}}$ & 0.11 & 6.30 & 4.90 & 17.15 & - \\
        & Ours & \textbf{0.05} & \textbf{2.80} & \textbf{42.92} & \textbf{67.90} & \textbf{0.89} \\
        \midrule
        \multirow{6}{*}{Dense Grid} & AD & 0.11 & 6.55 & 11.49 & 29.40 & - \\
        & FD & 0.07 & 3.97 & 30.66 & 55.06 & 2.62 \\
        & GAD & \textbf{0.05} & \textbf{3.24} & \textbf{40.50} & \textbf{64.01} & - \\
        & SFD & 0.94 & 57.62 & 0.01 & 0.07 & - \\
        & $\text{SFD}_{\text{256}}$ & 0.10 & 6.12 & 5.09 & 17.64 & - \\
        & Ours & \underline{0.06} & \underline{3.31} & \underline{38.95} & \underline{62.65} & \textbf{0.89} \\
        \midrule
        \multirow{6}{*}{Tri-plane \cite{Chan2021Triplane}} & AD & 0.15 & 8.59 & 3.61 & 13.13 & - \\
        & FD & 0.07 & 4.19 & 23.42 & 51.27 & 4.12 \\
        & GAD & \textbf{0.05} & \textbf{2.92} & \underline{34.75} & \textbf{64.23} & - \\
        & SFD & 0.94 & 57.65 & 0.01 & 0.07 & - \\
        & $\text{SFD}_{\text{256}}$ & 0.10 & 6.23 & 4.88 & 17.19 & - \\
        & Ours & \underline{0.06} & \underline{3.23} & \textbf{35.67} & \underline{62.74} & \textbf{0.90} \\
        \bottomrule
    \end{tabular}}
    \vspace{-0.7em}
    \caption{\textbf{Operator evaluation}. We compare our approach with the baselines on the FamousShape dataset \cite{ErlerEtAl:Points2Surf:ECCV:2020}. We report the performance averaged over the dataset. }
    \label{tab:opacc}
\vspace{-1.2em}
\end{table}

\begin{table}[h!]
    \centering
    \renewcommand{\arraystretch}{1.2}
    \renewcommand{\tabcolsep}{1.2mm}
    \resizebox{\linewidth}{!}{
    \begin{tabular}{cccccccc}
        \toprule
        \multirow{2}{1.5cm}{Model} & \multirow{2}{2.1cm}{Fine-tuning/ Regularization$^\star$ method} & \multicolumn{4}{c}{Autodiff Surface Normal} & \multicolumn{2}{c}{Mesh Reconstruction} \\
        \cmidrule(lr){3-6} \cmidrule(lr){7-8}
        & & L2 $\downarrow$ & Ang $\downarrow$ & AA@1 $\uparrow$ & AA@2 $\uparrow$ & CD $\downarrow$ & F-Score $\uparrow$ \\
        \midrule
        \multirow{5}{1.5cm}{Instant NGP \cite{ingp}} & - & 0.21 & 12.40 & 1.58 & 6.12 & $9.24 \times 10 ^{-4}$ & 93.07 \\
        & Eikonal$^\star$ & 0.11 & 6.51 & 12.24 & 31.48 & $9.23 \times 10 ^{-4}$ & 92.90 \\
        & FD-Eikonal$^\star$ & 0.20 & 12.46 & 0.48 & 6.04 & $9.20 \times 10 ^{-4}$ & 93.09 \\
        & FD & \underline{0.08} & \underline{5.14} & \underline{21.16} & \underline{46.63} & $9.35 \times 10 ^{-4}$ & 90.24 \\
        & Ours & \textbf{0.05} & \textbf{3.19} & \textbf{33.60} & \textbf{60.24} & $9.28 \times 10 ^{-4}$ & 92.28 \\
        \midrule
        \multirow{5}{1.5cm}{Dense Grid} & - & 0.11 & 6.56 & 11.42 & 29.37 & $9.26 \times 10 ^{-4}$ & 89.83 \\
        & Eikonal$^\star$ & 0.16 & 9.82 & 12.70 & 27.42 & $9.24 \times 10^{-4}$ & 87.79 \\
        & FD-Eikonal$^\star$ & 0.10 & 6.17 & 13.71 & 33.27 & $9.25 \times 10^{-4}$ & 89.85 \\
        & FD & \underline{0.09} & \underline{5.09} & \underline{18.82} & \underline{41.52} & $9.23 \times 10 ^{-4}$ & 88.94 \\
        & Ours & \textbf{0.08} & \textbf{4.40} & \textbf{29.32} & \textbf{51.40} & $9.25 \times 10 ^{-4}$ & 87.66 \\
        \bottomrule
    \end{tabular}}
    \vspace{-0.7em}
    \caption{\textbf{Effect of fine-tuning}. We compare autodiff operators before (first row) and after fine-tuning with different operators along with common regularization approaches ($\star$) for neural fields. The accuracy of autodiff surface normals improves after fine-tuning.}
    \label{tab:ftres}
\vspace{-1.5em}
\end{table}

\subsection{Improving pre-trained neural fields.}

 We next evaluate whether the fine-tuning approach proposed in Section \ref{sec:ft} improves the autodiff derivative estimates. Since our hybrid neural fields do not admit higher-order derivatives, we evaluate only the first-order derivatives.
 We evaluate two versions of our fine-tuning approach, one using finite difference-based gradient operators as supervision, and the other using our polynomial fit-based operator.
 We compare the autodiff gradients after fine-tuning to the un-finetuned network. We also compare with networks trained from scratch using the commonly used eikonal regularization~\cite{atzmon2019sal, gropp2020implicit} for neural fields, proposed to learn smooth iso-surfaces without disturbing the fidelity of the original neural field, including its finite differences-based variant (FD-Eikonal)~\cite{li2023neuralangelo}. 
 We only performed experiments on Instant NGP and Dense Grid as our Tri-plane implementation did not support higher-order derivatives.
 Our results (\Cref{tab:ftres}) demonstrate that fine-tuning improves derivative estimates significantly, with our polynomial fit-based operator providing better supervision. 
We also observe an improvement in gradient accuracy over the regularization approaches (marked $\star$).
Furthermore, the fine-tuning process preserves the zero-level set of the pre-trained hybrid neural field, as highlighted by minor changes in Chamfer Distance (CD) and F-Score. 

\section{Applications}
\label{sec:applications}

We now demonstrate the impact of our improved derivatives on downstream applications. For implementation details of the applications, see \conditionalLink{sec:appsetups}{Appendix D}.

\subsection{Rendering}
\label{sec:rendering}
In rendering, accurate surface normals (which correspond to the gradient of the SDF) are needed to estimate how light will reflect off a surface \cite{fog}.
We show the impact of our improved gradients on the rendering of a hybrid neural SDF representing a perfectly specular sphere, and another representing a perfectly lambertian Armadillo \cite{armadillo}.

For the sphere, we use the analytic SDF and surface normals for the ground truth, while we use a mesh as reference for the Armadillo.
The sphere was lit with an environment map, and the armadillo with a light source from behind the camera.
We use sphere tracing to compute the first ray intersection from the camera with the zero-level iso-surface. Subsequently, we queried the network to obtain gradients using automatic differentiation, finite differences, our post hoc polynomial-fitting operator, and autodiff gradients obtained from a network that was fine-tuned with our operator.

\begin{figure*}[!h]
    \centering
        \includegraphics[width=0.95\linewidth]{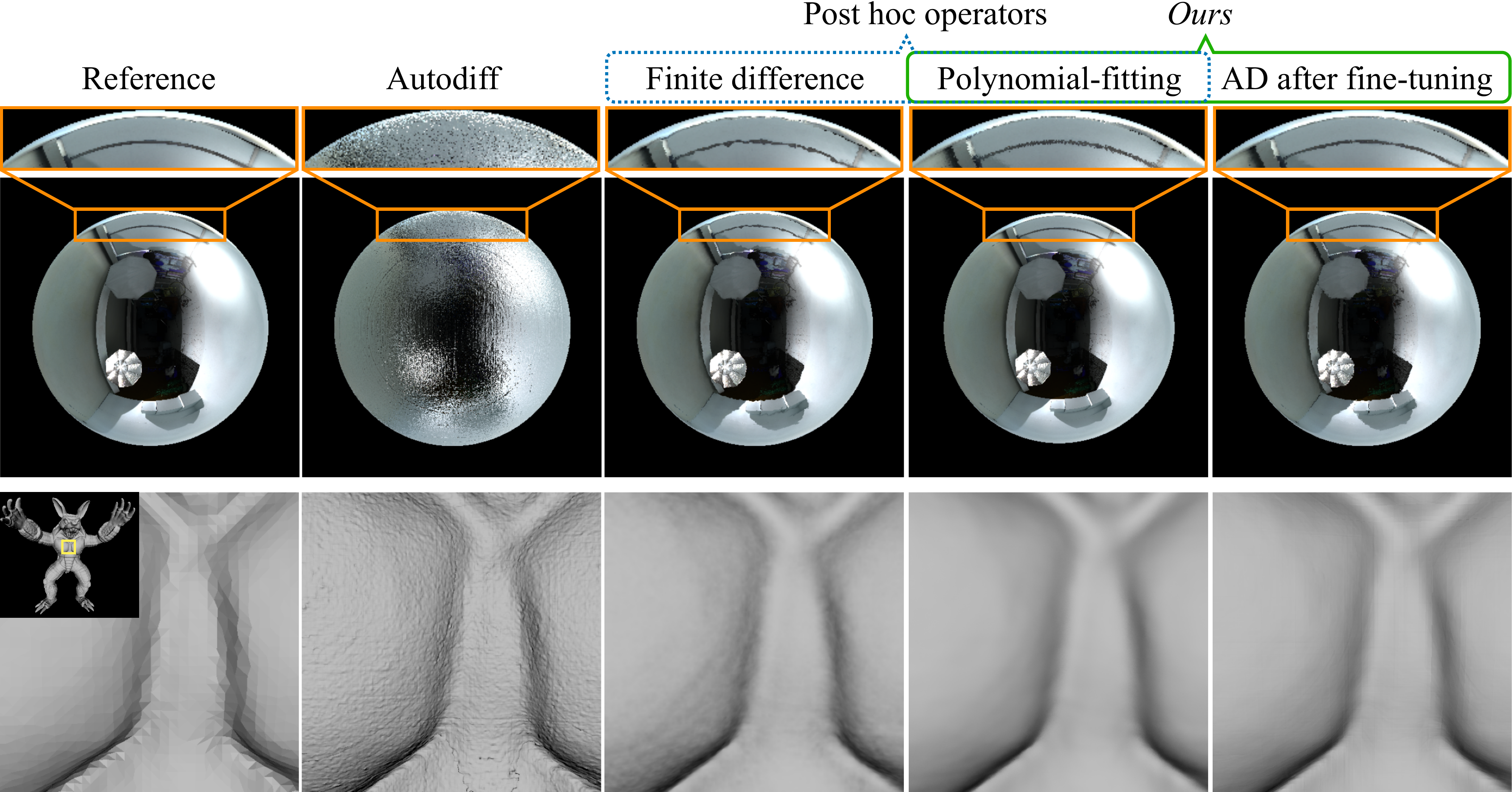}
    \vspace{-0.5em}
    \caption{\textbf{Accurate Normals for Rendering.} A perfectly specular sphere lighted by an environment map (top) and a diffuse Armadillo (inset) lit by a light source put in front of the object (bottom). In both cases, noisy normals from autodiff lead to artifacts in rendering as shown in the highlighted parts for the sphere and the chest of the Armadillo, that are mitigated by our approaches.}
    \label{fig:rendering}
    \vspace{-1.5em}
\end{figure*}

Figure \ref{fig:rendering} presents our results. As predicted, for the supposedly smooth sphere, as well as the Armadillo, we observed severe surface artifacts using gradients from autodiff. The finite difference-based post hoc operator is able to tackle noise to an extent but still leads to artifacts. On the other hand, normals estimated by our approaches give a much more noise-free image that closely matches the reference. We also provide additional results in \conditionalLink{sec:addlrendering}{Appendix G}. 

\subsection{Simulating Collisions}
\label{sec:collision}
When simulating collisions between objects, normals help determine the impulse direction~\cite{Catto2005IterativeDW,erleben}. 
When working with hybrid neural SDFs, we would need to query the normal at the local coordinates of the point of collision to the network. 
If the normals are inaccurate, this can lead to incorrect object trajectories after the collision.

\begin{figure}[t]
    \centering
        \includegraphics[width=0.49\linewidth]{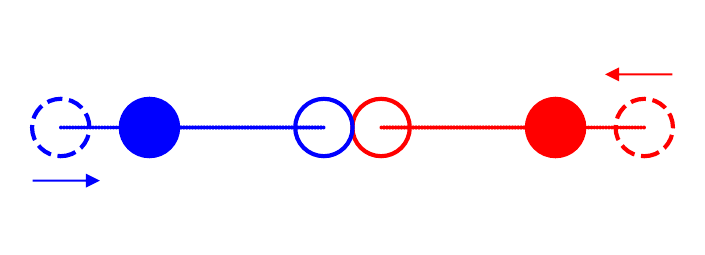}
        \includegraphics[width=0.49\linewidth]{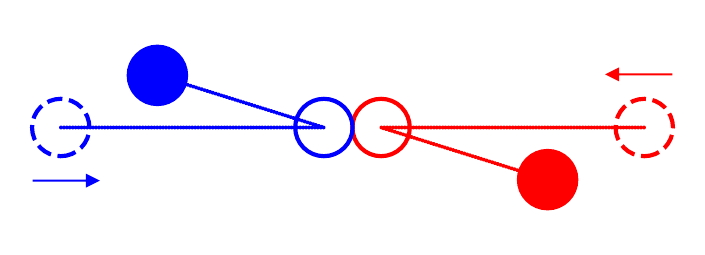}
    \vspace{-1em}
    \caption{\textbf{Illustration of how noisy normals affect collision.} Two spheres undergoing perfectly elastic head-on collisions simulated using correct surface normals will re-trace their paths after a collision. However, inaccurate normal estimates from autodiff yield incorrect trajectories after bouncing (right).}
    \label{fig:collision}
    \vspace{-1em}
\end{figure}

For our experimental setup, we consider two identical spheres undergoing head-on collision on a plane and simulate their trajectories post-collision.
To obtain these trajectories, we use the normal estimates from the two hybrid neural SDFs at the point of contact.
We model the collisions as perfectly elastic so that there is no loss of energy.
Ideally, the spheres should rebound along the line joining the centers, but inaccurate normals will lead to incorrect trajectories. \Cref{fig:collision} illustrates such a simulation and also shows how things fail when using autodiff to compute normals. Averaged over $10^6$ trials, the error obtained from our normals was $\mathbf{0.85^\circ}$, compared to $\mathbf{11.51^\circ}$ for autodiff normals.

\subsection{PDE Simulation}
\label{sec:pde}
\vspace{-0.2em}
Recently, Chen~\etal~\cite{chenwu2023insr-pde} proposed using Implicit Neural Spatial Representations (INSR) as the spatial representation of the PDE solution instead of explicit spatial discretization. We build upon their work and highlight that accurate gradient operators also enable the use of hybrid neural fields for PDE simulation.
We simulate a 2D advection equation,$
    \frac{\partial u}{\partial t} = - a \nabla_\mathbf{x} u
$.
For the initial condition, we use a Gaussian pulse centered at $(-0.6, -0.6)$ with a standard deviation of $0.1$. We choose a constant velocity, $a = [0.25\ 0.25]^T$. We run our simulations in a square of side length 2 centered at (1, 1). We use the Dirichlet boundary condition, i.e., the field becomes 0 at the boundary, same as INSR \cite{chenwu2023insr-pde}. For time integration, we use the forward Euler method, given by,
$
    u^{t+1} = u^t - a\Delta t\nabla_\mathbf{x}u
$.
\begin{figure}[t]
    \centering
    \includegraphics[width=0.9\linewidth]{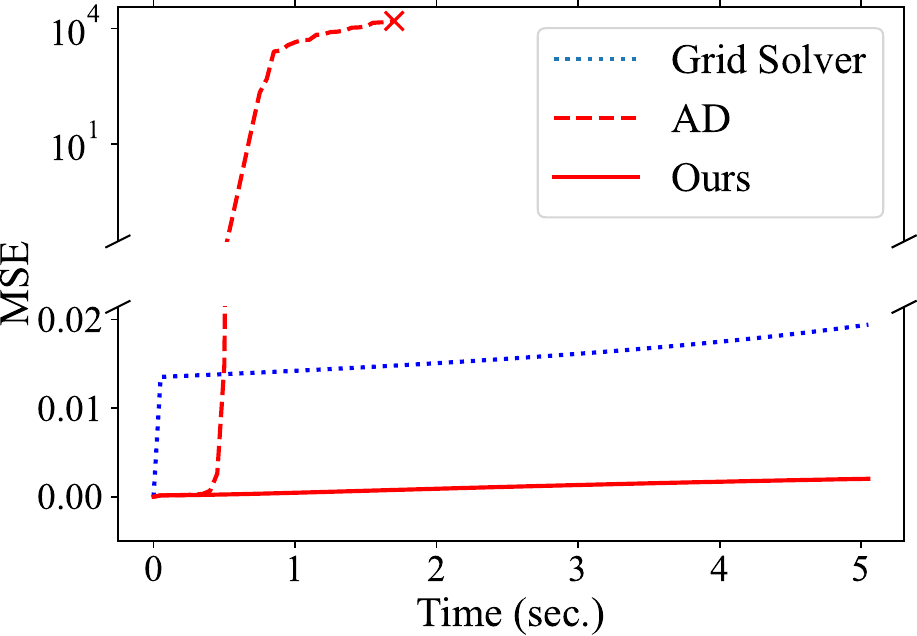}
    \vspace{-0.8em}
    \caption{\textbf{Effect of inaccurate gradients in PDE simulation}. Mean squared error (MSE) for 2D advection for a finite difference grid solver, autodiff gradients (AD), and our polynomial-fitting approach. Error for AD explodes after the first few seconds and eventually crashes (indicated by $\times$).}
    \label{fig:pdesim}
    \vspace{-1.5em}
\end{figure}

While INSR uses a non-hybrid neural field (SIREN~\cite{sitzmann2020siren}) for representing the PDE solution, we use a hybrid neural field. In our setup, the gradient of the initial condition ($\nabla_\mathbf{x}u$) can either be queried using autodiff or using our operator. For evaluation, we compare the error (w.r.t. the analytical solution) in the evolution using our polynomial-fitting gradient operator with autodiff (AD) gradients\footnotemark. We also show the error from a finite difference-based grid solver to show where traditional methods stand. All the methods use a step size ($\Delta t$) of $0.05$, and we run our solvers for $100$ time steps. 
\Cref{fig:pdesim} shows our results. The grid solver accumulates errors over time due to numerical dissipation caused by its spatial discretization. Using hybrid neural fields with autodiff gradients leads to diverging solutions and the evolution collapses after 2 seconds. 
Using the same hybrid neural field with our operator leads to more accurate solutions 
at all time steps.
\footnotetext{While we cannot make an apples-to-apples comparison with INSR as they use a non-hybrid neural field, for completeness, we provide a comparison in \conditionalLink{sec:addlresultsinsr}{Appendix B.4}.}

\section{Limitations and future work}
One limitation of our approach is the need to set the hyper-parameter $\sigma$ based on the downstream application. 
However, note that analogous hyperparameters are common in other related problems where smoothing is required: \eg, derivative computation in image processing or fitting surfaces to point clouds with MLS~\cite{2004shortmlsintro}. One may argue that in these methods and in our approach, the ability to set $\sigma$ offers an additional degree of control.

A second limitation is that our approach needs to sample the neighborhood of the query point, necessitating several forward passes per query (although we observed that our operator performs competitively with alternatives like finite differences, see \conditionalLink{sec:addlresultsrt}{Appendix B.3}).
This cost may be amortized by our fine-tuning approach.
Alternatively, clever sharing of samples between neighboring query points is an interesting avenue for future work.

Finally, our approach is primarily designed to remove high frequency noise.
As such, it cannot help remove other, more lower frequency errors that are common in non-hybrid neural field architectures such as SIREN~\cite{sitzmann2020siren} (\conditionalLink{app:nonhybrid}{Appendix E}).

\section{Conclusion}
In this paper, we have shown that automatic differentiation of trained hybrid neural fields yields extremely noisy derivatives and impacts several downstream applications.
We tackle this problem with a new derivative operator that computes the derivative on a local polynomial approximation of the hybrid neural field.
We further propose a self-supervised fine-tuning approach to improve the accuracy of autodiff gradients directly.
We demonstrate significant improvements in derivative accuracy from these new techniques.
We further demonstrate that our methods improve performance in rendering and physics simulation applications compared to directly using autodiff derivatives for hybrid neural fields.

\section*{Acknowledgements}

This work was partly funded by NSF IIS: 2144117, NSF IIS: 2107161 and NSF HCC: 2212084.   We would like to thank Peter  Michael for help with the initial implementation of the rendering experiments, Yihong Sun for help with some of the figures, Gemmechu Hassena for providing meshes for some of the rendering results, and Mariia Soroka for discussions on Monte Carlo derivative estimation.

{
    \small
    \bibliographystyle{ieeenat_fullname}
    \bibliography{egbib,references/rw_rendering, references/nf}

}
\maketitlesupplementary
\appendix

\section{Experimental details}
\label{sec:expdetails}
In this section, we provide the implementation details for our experiments described in \conditionalLink{sec:experiments}{Section 4}.

\subsection{Dataset} We perform pre-training on shapes from the FamousShape dataset \cite{ErlerEtAl:Points2Surf:ECCV:2020}. We filter out shapes with non-watertight meshes or incorrectly oriented normals. This is because non-watertight meshes do not admit a valid SDF and in order to compute the correct ground truth, we require meshes with correct normals. This gave us a set of 15 shapes. We further center the meshes at the origin and normalize them to lie inside the $[-1, 1]^3$ cube.

\subsection{Pre-training} 
\label{sec:pretraining}

The inputs for our experiments are the pre-trained hybrid neural SDFs of the shapes. In this section, we present details about how we obtain the pre-trained models. First, we provide a description and architectural details of our hybrid neural fields:
\begin{itemize}
    \item Instant NGP \cite{ingp}: We retained the original architecture from the paper. We implemented our models using \texttt{tiny-cuda-nn} \cite{tiny-cuda-nn}.
    \item Dense Grid: A grid-based neural field with dense feature grids, as discussed in M\"uller~\etal~\cite{ingp}. We use a multi-resolution grid with 4 levels, starting from a minimum resolution of 16 up to a maximum resolution of 256.
    \item Tri-Plane \cite{Chan2021Triplane}: Instead of volumetric grids, they consist of 3 planar grids (one each for XY, YZ, and XZ planes), with a feature embedding residing on each grid point. For a given query point, the features are combined using bi-linear interpolation on each plane and then further summed together. Finally, the feature is passed through an MLP to obtain the output. We used planes with a resolution of 512, feature embeddings of size 32, and an MLP with 2 hidden layers of size 128.
\end{itemize}
  We follow the same data sampling procedure for training neural SDFs as described by Müller~\etal~\cite{ingp} for training the Instant NGP.  We trained all models for $10^4$ steps using the Adam \cite{kingma2014adam} optimizer with an initial learning rate of $1\mathrm{e}{-3}$ and reduced the learning rate by a factor of 0.2 every 5 steps. 

\subsection{Post hoc operator}
\label{sec:posthocdetails}
In this section, we provide details for the hyperparameter selection procedure used for our post hoc polynomial-fitting operator. We used a fixed value of 256 for $k$. For the value of $\sigma$, we selected the best value using telescopic search in two levels: the first sweep is conducted over ${10^i: -5 \leq i \leq 1}$, after which we zoom in to the interval bounded by the best value, $\sigma_1$ and its best neighbor $\sigma_2$. Assuming here for simplicity that $\sigma_1 < \sigma_2$, we then conduct a sweep over 20 values taken at uniform intervals from $[\sigma_1, \sigma_2]$. 
\vspace{-1.2em}
\paragraph{Baselines.} We compare our polynomial-fitting operator with automatic differentiation and finite difference for computing surface normals and mean curvatures of the shapes. For automatic differentiation, we directly query the network using PyTorch's \cite{paszke2019pytorch} automatic differentiation toolkit. For the finite difference operators, we used a centered difference approach, sampling local axis-aligned neighbors of the query point and using them to compute the operator. The finite difference operator had a hyperparameter $h$ for the stencil size. In essence, it gives the size of the finite difference grid cell, if we were to set up a global grid for computing finite differences
. We selected this hyperparameter by sweeping over the set $\{\frac{2}{2^i}: 5 \leq i \leq 9\}$. Here $2^i$ is analogous to the resolution of the global finite difference grid. 

\subsection{Fine-tuning}
\label{sec:ftdetails}

As discussed in \conditionalLink{sec:ft}{Section 3.3}, we train an ensemble of models where each model is supervised with a different version of the smoothed gradient operator, $\hat{\nabla}_\mathbf{x}$ characterized by the amount of smoothing it imposes. For fine-tuning based on polynomial-fitting derivatives, we ensemble using $\sigma$ values taken uniformly from the interval $[1\mathrm{e}-3, 1\mathrm{e}-2]$ at steps of $5\mathrm{e}-3$. For finite-difference-based fine-tuning we ensemble using stencil sizes from the set $\{2^i: 5 \leq i \leq 9\}$. We fine-tune all models for 4000 steps with a constant learning rate of $2\mathrm{e}-3$, using the Adam optimizer \cite{kingma2014adam}. 

\vspace{-1em}
\paragraph{Baselines.} We have also compared our fine-tuning approaches with neural fields trained using eikonal regularization~\cite{atzmon2019sal, gropp2020implicit}. For the commonly-used eikonal loss that uses autodiff gradients, we follow the same training parameters as previously described in pre-training (\Cref{sec:pretraining}). We just add the eikonal loss to the loss function with a weight of $10^{-3}$ (selected by sweeping over $\{1, 10^{-1}, 10^{-3}\}$). For the finite differences-based variant of the eikonal loss~\cite{li2023neuralangelo}, we found that the same weight gave the best result, and for the size of the finite-difference stencil ($\epsilon$ in Li \etal) we selected the value for each shape by sweeping over the set $\{\frac{2}{2^i}: 5 \leq i \leq 10\}$.

\subsection{Evaluation}

To compare our approach against the baselines, we generate ground truth surface normals and mean curvatures using the meshes of the shapes. First, we compute the vertex normals and discrete mean curvatures \cite{discretecurvature} of the shapes from the meshes. Next, we sample $2^{18}$ points on the surface of the meshes. We interpolate the vertex normals and mean curvatures to each point using barycentric interpolation from the mesh vertices. This set of points, their computed normals, and mean curvatures become the ground truth used in our evaluations. The metrics used for our evaluations have been described in \conditionalLink{sec:experiments}{Section 4} (under \textit{Metrics}).

\section{Additional Results}
\label{sec:addlresuls}
\subsection{Accuracy analysis}
\label{sec:addlaccanalysis}

In \conditionalLink{sec:experiments}{Section 4}, we reported the results for the accuracy of our operators. In this section, we provide the full results for the accuracy analysis of our operators and our fine-tuning approach on the FamousShape dataset \cite{ErlerEtAl:Points2Surf:ECCV:2020}. \Cref{tab:opaccfull} shows comparisons between our post hoc operator and the baselines on Instant NGP while \Cref{tab:ftresfull} shows how our best fine-tuning approach, i.e., fine-tuning with polynomial-fitting gradients.
We show our results for Dense Grid models in \Cref{tab:opaccdense,tab:ftresdense}. We can observe that we obtain more accurate gradients than the baselines. This also shows that the artifacts that we observed in the case of Instant NGP were not solely a result of its hash encoding.
Finally, results presented in \Cref{tab:opacctri} show that even on a significantly different hybrid architecture like Tri-plane, our operators can provide more accurate surface normals and mean curvatures. At the time of writing, our Tri-plane implementation did not have support for higher-order derivatives through autodiff derivatives. Hence, we were unable to show fine-tuning results.

\subsection{Results on images} 
\label{sec:addlresultsimages}

\begin{figure*}
    \centering
    \includegraphics[width=0.9\linewidth]{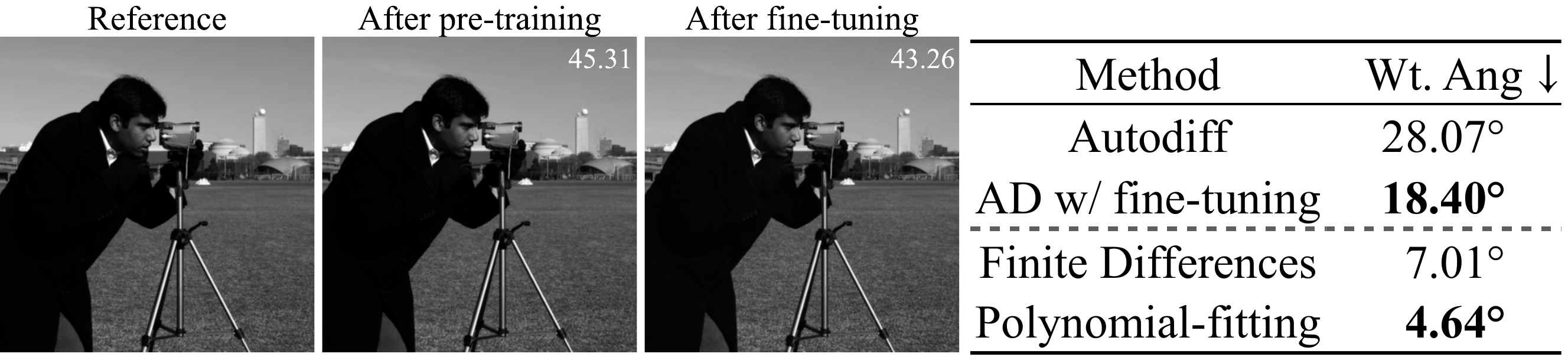}
    \caption{\textbf{Results on images.} We show the application of our operators on a hybrid neural field trained to represent an image. For reference, we use the image derivative obtained using Sobel filtering, similar to Sitzmann~\etal~\cite{sitzmann2020siren}. We compare the image gradient obtained using our post hoc and fine-tuning approaches with the baselines. For the zeroth-order signal, we show the PSNR (inset) which shows that fine-tuning preserves the initial image. For the image gradient, we show the weighted mean angular error, weighted by the reference gradient magnitude. Applying autodiff after our fine-tuning approach leads to more accurate gradients than direct autodiff. Using our post hoc operator also leads to more accurate gradients than finite differences.}
    \label{fig:imgresults}
    \vspace{-1em}
\end{figure*}

We also show the benefits of our approaches on a different modality, specifically images. We train an Instant NGP \cite{ingp} model on an image and evaluate its derivatives using our proposed approaches. For pre-training our model, we used a relative L2 loss and trained using the Adam optimizer with a learning rate of 0.01. For fine-tuning, we use MSE loss for $\mathcal{L}_{\mathrm{con}}$, and weighted weighted $\mathcal{L}_{\mathrm{grad}}$ by $10^{-3}$, and trained using a learning rate of 0.02.

 \Cref{fig:imgresults} shows our results. For reference, we use the derivatives obtained using Sobel filtering, similar to Sitzmann~\etal~\cite{sitzmann2020siren}. Firstly, we observe that our fine-tuning approach preserves the initial image, with a minor drop in the PSNR over the pre-trained image. We also compare the accuracy of derivatives using a weighted mean angular error, where the weights are the reference gradient magnitudes. This is because image gradients are usually more important in regions with high gradient magnitudes (the edges). Our post hoc operator gives more accurate gradients than finite differences. We also observe that autodiff gradients obtained after our fine-tuning approach are more accurate than naively applying autodiff to the pre-trained signal.

\subsection{Runtime Analysis}
\label{sec:addlresultsrt}
We compare the wall time of our local polynomial-fitting approach with finite difference and autodiff operators. For our operator, we use $k=256$. We computed the mean and standard deviation of wall-time required by all methods on a single query point, averaged over 7 runs each running 1000 instances of the method. 

\begin{wraptable}{r}{0.2\textwidth}
\small
    \vspace{-2.5em}
    \centering
    \begin{tabular}{lc}
    \toprule
     Method    &  Time ($\mu$s)\\
    \midrule
     AD    & $1520 \pm 12.9$\\
     FD     &  $509 \pm 91.1$\\
     Ours & $459 \pm 16.3$\\
    \bottomrule\\
    \end{tabular}
\vspace{-1.5em}
\small
\caption{\textbf{Runtime Analysis}}
\label{tab:runtime}
\vspace{-1.75em}
\end{wraptable}

\Cref{tab:runtime} summarizes the results.
We found that our operator performs competitively in terms of runtime compared to finite difference (FD) and autodiff (AD) gradient operators. 
All these methods were benchmarked using an Instant NGP model \cite{ingp}. 

Our proposed fine-tuning approach takes $\sim$700s to reach $\sim$90\% of the reported performance.
Although vanilla Instant NGP can reach equivalent reconstruction loss in $\sim$20s, its derivatives are nowhere near as accurate as our approach even after $\sim$1000s worth of training.
That said, if training cost is a concern, we can trade off training cost for test-time compute using our post hoc operator. 
Also, note that ours is a naive implementation which can be sped up with engineering tricks (e.g., sharing local neighborhoods or sampling fewer points, trading off derivative accuracy).

\subsection{Comparing PDE simulation with INSR \cite{chenwu2023insr-pde}} 
\label{sec:addlresultsinsr}
While we have used the framework of INSR for our PDE simulation experiments, a direct apples-to-apples comparison with INSR is not possible due to INSR utilizing a different architecture (SIREN). As we discuss later (in Appendix \ref{app:nonhybrid}), while SIREN also suffers from inaccurate derivatives, the nature and cause of those accuracies differ significantly from the high-frequency noise that we claim to address. Tackling SIREN's derivative errors would require an altogether different approach that we hope to address in future work.

\begin{figure}[t]
    \centering
    \includegraphics[width=\linewidth]{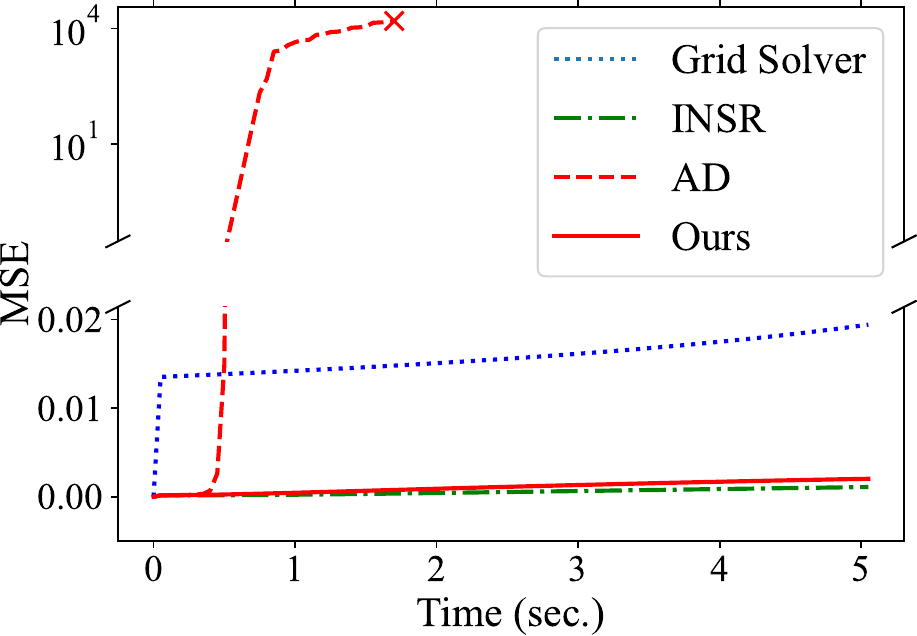}
    \caption{\textbf{Comparison to INSR \cite{chenwu2023insr-pde}}. While INSR performs better, our approach allows hybrid neural fields to perform competitively, which is not possible when using autodiff gradients directly.}
    \label{fig:pdesiminsr}
    \vspace{-1em}
\end{figure}

However, to show how our approach with hybrid neural fields stands relative to a current state-of-the-art approach like INSR, we show a comparison between the errors of our approach and INSR in the same setup as discussed in \conditionalLink{sec:pde}{Section 5.3}. Figure \ref{fig:pdesiminsr} shows our results. We can see that while INSR performs better, using our approach to compute derivatives with hybrid neural fields allows hybrid neural fields to perform competitively against INSR, which is not possible with autodiff derivatives.

\section{Comparison to Marching Cubes} 
\label{sec:mccmpr}
\begin{figure*}
    \centering
    \includegraphics[width=0.8\textwidth]{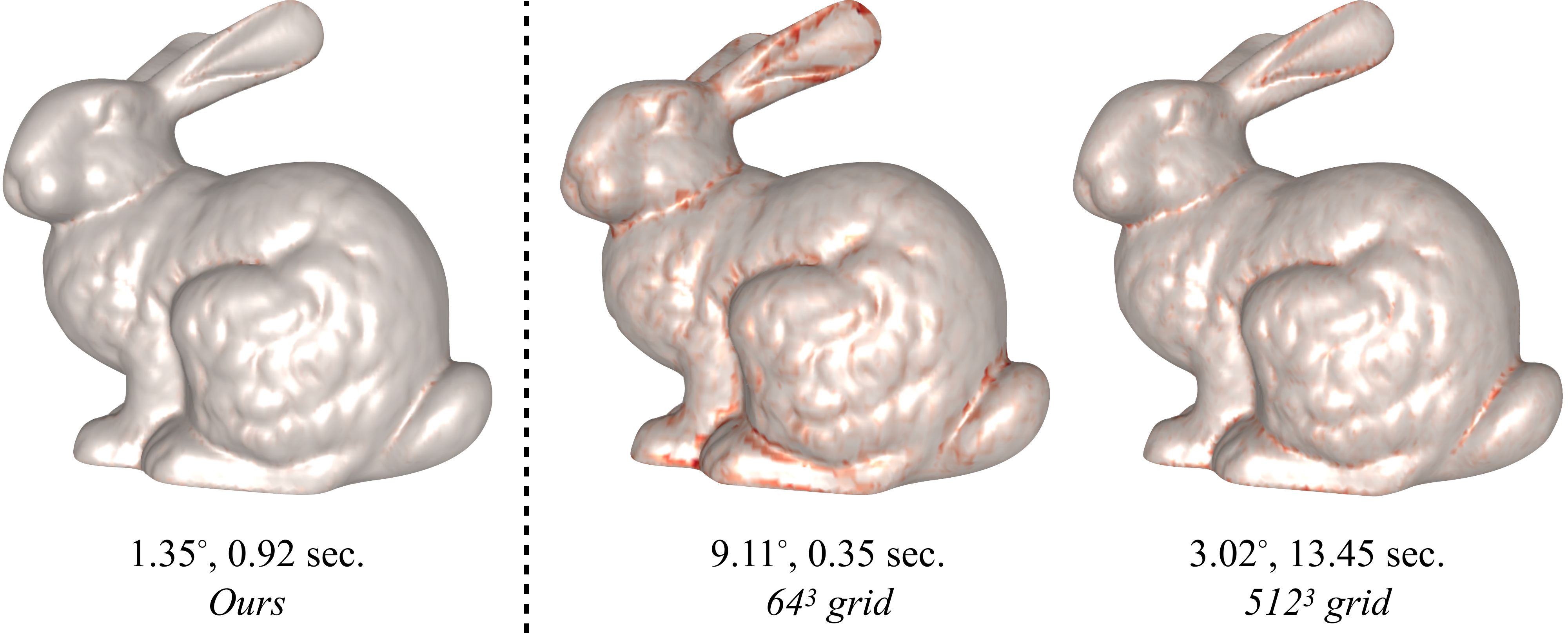}
    \caption{\textbf{Marching Cubes for derivatives.} Mean angle error and time required by Marching Cubes to compute surface normals (first-order derivative) on the Bunny shape (Red denotes error). This approach can be expensive ($\mathbf{15\times}$ time) for obtaining accurate surface normals. Reducing the grid resolution can reduce time but trades off accuracy for efficiency ($\mathbf{7\times}$ error). Comparatively, our approach provides accurate normals efficiently.}
    \label{fig:mcreason}
    \vspace{-1.5em}
\end{figure*}

As discussed in \conditionalLink{sec:localpolyfit}{Section 3.2}, one other alternative for computing derivatives is by directly extracting the mesh using the Marching Cubes algorithm \cite{Lorensen1987MarchingCubes}. While mesh extraction with Marching Cubes can take time, this cost can be amortized over multiple queries for derivatives using the extracted mesh. Hence, for a fair runtime comparison to Marching Cubes, we compare the runtime of our operator with Marching Cubes on a larger point set of size $2^{18}$ sampled uniformly from a 3D shape, in this case, the Stanford Bunny. Since the points sampled may not always lie on the extracted mesh for Marching Cubes, we compute the normals at the closest on-surface point. Figure \ref{fig:mcreason} illustrates how getting comparably accurate derivatives requires running Marching Cubes at a high grid resolution (512) which takes up almost $\mathbf{15\times}$ the time taken by our approach. We can try to save time by running marching cubes at a lower resolution, however, this leads to inaccurate derivatives, resulting in almost $\mathbf{7\times}$ the error incurred by our approach. Thus, getting accurate derivatives from Marching Cubes is quite expensive compared to our approach, and can become increasingly prohibitive in applications like physical simulation, where frequent derivative queries may be required from an evolving underlying signal.

\section{Application Setups}
\label{sec:appsetups}
In this section, we describe the details of the experiential setup used in each application described in \conditionalLink{sec:applications}{Section 5}.

\paragraph{Rendering.} In our rendering experiments (\conditionalLink{sec:rendering}{Section 5.1}) for both shapes, we used the Instant NGP model \cite{ingp}. The training and hyperparameter selection were done using the same process as described in \Cref{sec:pretraining} and \Cref{sec:ftdetails} respectively. For our polynomial-fitting operator, we use $\sigma=0.03$ and $k=256$ for the sphere and $\sigma=0.002$ and $k=256$ for the Armadillo, selected using telescopic search. For the results of the fine-tuning approach, we queried all models in the ensemble and selected the best render after visual comparison. For the finite difference operator, we selected a stencil size of $\frac{2}{32}$ for the sphere and $\frac{2}{512}$ by conducting a sweep as described in \Cref{sec:posthocdetails}.

\paragraph{Simulating Collisions.} For our experiments on simulating collisions (\conditionalLink{sec:collision}{Section 5.2}), the hybrid neural SDF of the sphere was a Dense Grid model. It had a minimum resolution of 16, a maximum resolution of 128, and consisted of 4 grid levels. For our polynomial-fitting operator, we used $\sigma=0.03, k=64$, selected using telescopic search.

\paragraph{PDE simulation.} For the PDE simulation experiment (\conditionalLink{sec:pde}{Section 5.3}), we used the same model architecture as the collision experiments, with a minimum resolution of 16, a maximum resolution of 128, and 4 grid levels. We modify the code shared by the authors of INSR \cite{chenwu2023insr-pde} to solve the 2D advection problem. However, we retain the data sampling and the training strategies used by the authors such as uniform sampling of the domain for training the implicit field, and early stopping during optimization. Our initial condition is a Gaussian pulse, given by:

\begin{equation}
    f(x, y) = e^{-\big (\frac{(x - \mu_1)^2 + (y - \mu_2)^2}{2 \sigma ^2}\big)}
\end{equation}

where $\mu_1 = -0.6, \mu_2 = -0.6, \sigma=0.1$. We run our simulations in a square of side length 2 centered at (1, 1). For the boundary conditions, we use the Dirichlet boundary condition, i.e., the field becomes 0 at the boundary, the same as INSR \cite{chenwu2023insr-pde} in their 1D advection setting. Other details are shared in Section \conditionalLink{sec:pde}{Section 5.3}.

\section{Effectiveness on a non-hybrid neural field (SIREN \cite{sitzmann2020siren})}
\label{app:nonhybrid}
\begin{figure*}
    \centering
    \begin{subfigure}[b]{0.32\linewidth}
         \includegraphics[width=\linewidth]{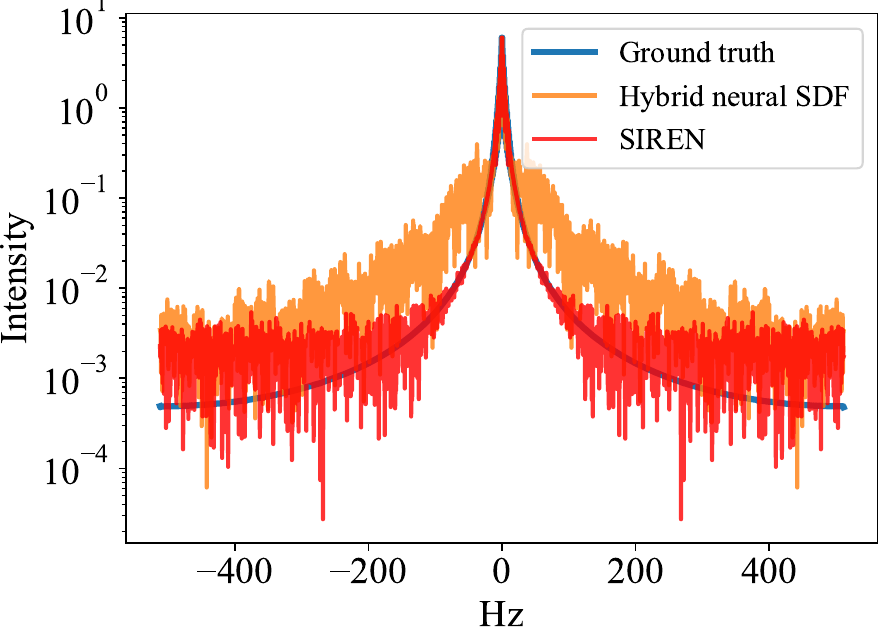}
         \caption{}
    \end{subfigure}
    \begin{subfigure}[b]{0.32\linewidth}
         \includegraphics[width=\linewidth]{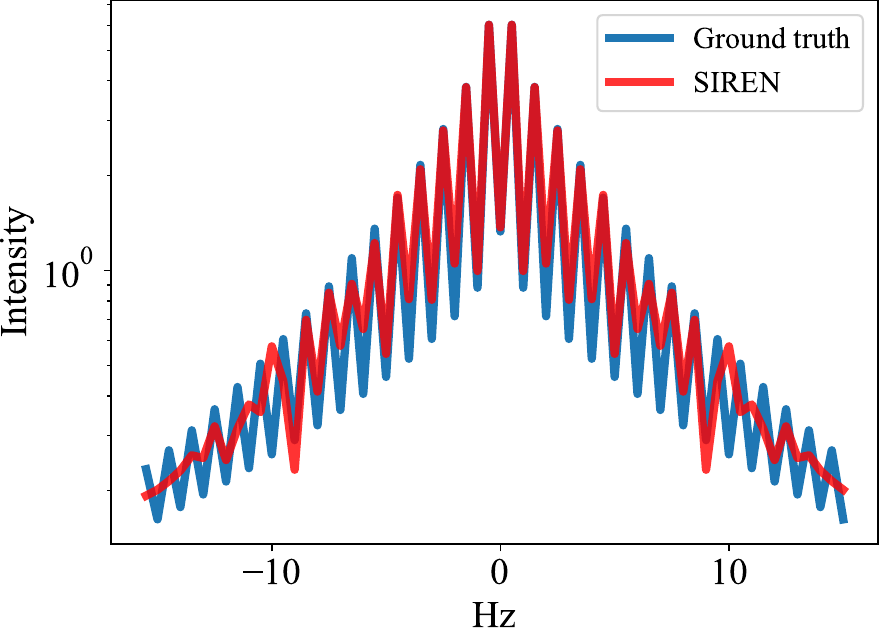}
         \caption{}
         \label{fig:zoominsiren}
    \end{subfigure}
    \begin{subfigure}[b]{0.32\linewidth}
         \includegraphics[width=\linewidth]{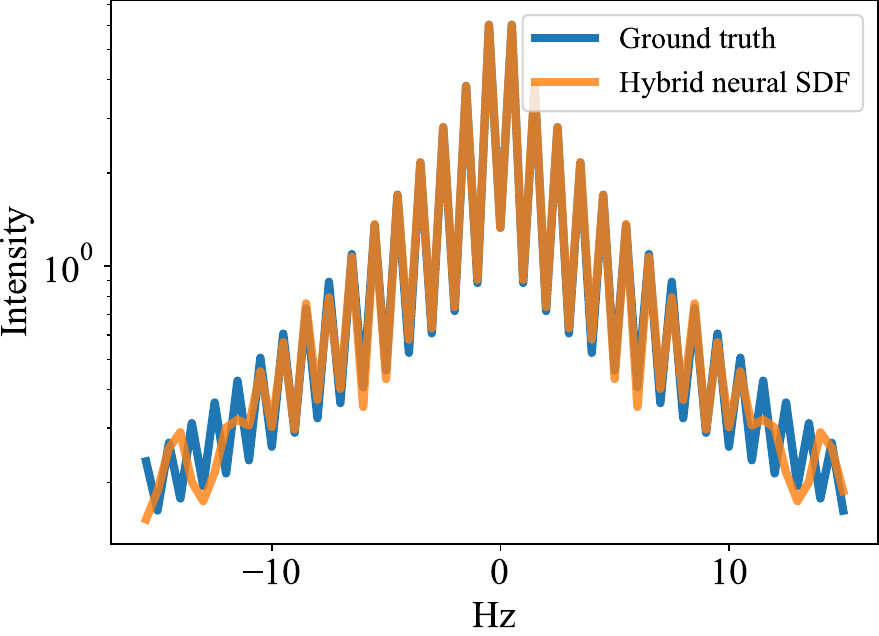}
         \caption{}
         \label{fig:zoominingp}
    \end{subfigure}
    \vspace{-1em}
    \caption{\textbf{Fourier spectrum of SIREN Vs. hybrid neural SDF.} Computed over a 1D slice (shown in Figure \ref{fig:sirenops}) of the SDF of a 2D circle. Note the lower degree of high-frequency noise compared to the hybrid neural SDF. Further zooming in (\Cref{fig:zoominsiren,fig:zoominingp}) to visualize the low-frequency components reveals the low-frequency errors in SIREN. Comparatively, the hybrid neural SDF more accurately captures the lower frequencies.}
    \label{fig:sirenfft}
    \vspace{-1em}
\end{figure*}

While our approaches are not tied to a particular architecture, they can only address the high-frequency noise in neural fields. As we illustrated in \conditionalLink{sec:method}{Section 3}, signals learned by hybrid neural fields like Instant NGP~\cite{ingp} are abundant in such high-frequency noise. 

We also investigated if similar kinds of artifacts arise in non-hybrid networks, specifically SIREN \cite{sitzmann2020siren}. We trained a SIREN network with $\omega_0=30$ and two hidden layers of size 128 each. Our first observation was that even for SIREN, derivatives, particularly higher-order derivatives, suffer from inaccuracies. However, unlike hybrid neural fields, we found that SIREN has a lower degree of high-frequency noise. The errors in SIREN seem to stem from low-frequency errors. \Cref{fig:sirenfft} illustrates this phenomenon. We observed similar trends for different values of $\omega_0$ $(\omega_0 \in \{20, 50\})$ and with varying hidden sizes (over $\{64, 256\}$)

\begin{figure*}
    \centering
    \includegraphics[width=0.8\linewidth]{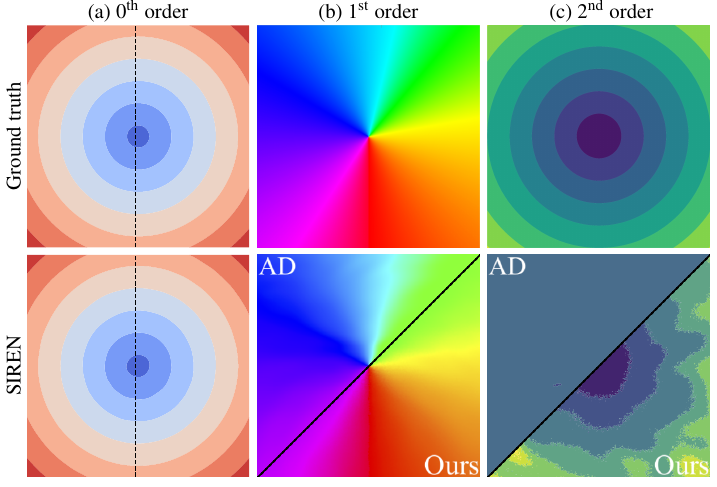}
    \vspace{-0.5em}
    \caption{\textbf{Differential operators of SIREN.} SIREN trained on the SDF of a circle in 2D. While the first-order operator (spatial gradient) for SIREN is quite accurate, the second-order operator (or the Laplacian) exhibits large errors. Applying our operators shows limited effectiveness, addressing the high-frequency noise in the signal but struggling with the low-frequency errors.}
    \label{fig:sirenops}
    \vspace{-1.5em}
\end{figure*}

Using our operators to compute the spatial derivatives of SIREN only helps to a limited degree (\Cref{fig:sirenops}). The observations on gradient are not very interesting as the autodiff gradient itself for SIREN is quite good and our operator leads to minor improvements. However, when computing the curvature (Laplacian), we observe that while autodiff curvatures are quite inaccurate, our operator can recover some reasonable values from the field, but noticeable errors remain. We believe that while our operator can address the high-frequency noise component in the underlying field, it is not able to overcome the low-frequency errors in SIREN.

To conclude, our preliminary experiments reveal that neural fields learned by SIREN have a lower degree of high-frequency noise and higher low-frequency errors compared to hybrid neural fields. As a result, while our operators can handle high-frequency noise, low-frequency errors still lead to inaccurate derivatives. Dealing with these low-frequency errors would require an altogether different approach and would be an interesting direction for future work.

\section{Training hybrid neural fields with accurate autodiff normals from scratch}
\label{app:latereg}

As discussed in \conditionalLink{sec:ft}{Section 3.3}, we can also use our proposed loss (\conditionalLink{eq:ftloss}{Eq. (3)}) for training hybrid neural fields from scratch. For this, we first train the model, $F_\Theta$ for $s\ (>0)$ steps as a \textit{warm-start} phase. This allows the model to learn a good initial estimate of the zeroth-order signal. Next, we train using our loss (\conditionalLink{eq:ftloss}{Eq. (3)}) for $(n - s)$ steps. For computing the smoothed gradient operator, we require $M$ which is a hybrid neural field with a good initial fit over the zeroth-order signal. In this case, we set $M$ as the frozen weights of $F_\Theta$ at the end of $s$ training steps. 

Intuitively, if $M$ fits the zeroth-order signal well, the smoothed gradient operator would be more accurate, leading to a more accurate supervision signal for $\mathcal{L}_\mathrm{grad}$. 
We want $s$ to be large enough so that we have a reasonably good fir with $M$.
Selecting a very small $s$ can lead to a poor fit and unstable optimization in the next stage. On the other hand, choosing a very large $s$ can lead to a time and resource-intensive training run. 
\begin{figure*}
    \centering
    \includegraphics[width=0.8\linewidth]{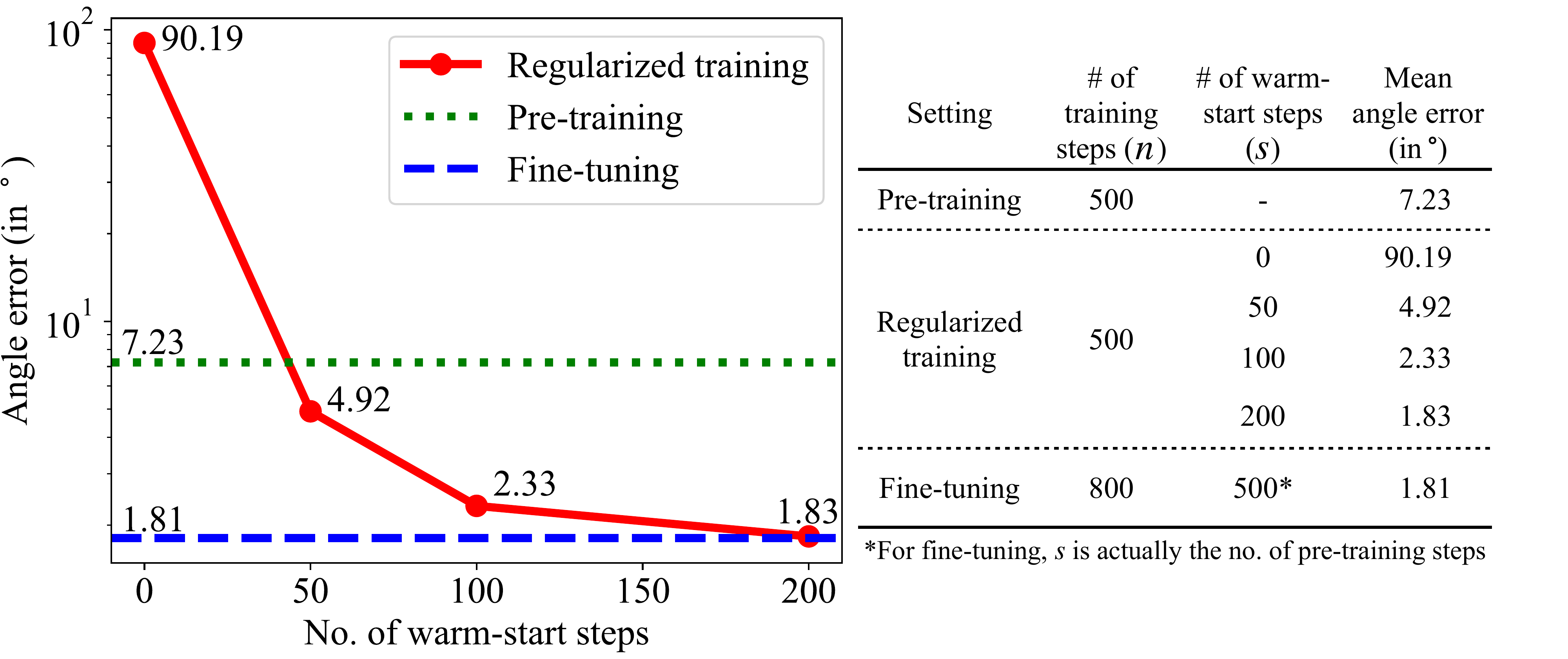}
    \caption{\textbf{Effect of $s$ on angle error.} We show the effects of the number of warm-start steps ($s$) on the accuracy of autodiff normals. We can observe that a higher value of $s$ leads to more accurate autodiff normals. We use the same Instant NGP architecture trained on the Armadillo shape for all settings. All models except for the fine-tuned version are trained for a total of 500 steps. The fine-tuned model (blue dashed) is trained for 300 more steps with our loss function after pre-training.}
    \label{fig:latereg}
    \vspace{-1em}
\end{figure*}

In this section, we analyze how the choice of $s$ affects the accuracy of autodiff normals. We train an Instant NGP~\cite{ingp} model on the Armadillo shape. We fix a total training budget of 500 steps. For each $s \in \{0, 50, 100, 200\}$, we train another hybrid neural field using the regularization approach described above and compare the angle error of their autodiff normals. \Cref{fig:latereg} shows our results. We also compare against a hybrid neural field that is trained normally, i.e., using only MSE loss for 500 steps. Let us consider this as the \textit{pre-trained} model (green dotted). We also fine-tune the pre-trained model using our fine-tuning approach described in \conditionalLink{sec:ft}{Section 3.3} for 300 more training steps (blue dashed). As expected, higher values of $s$ lead to more accurate autodiff normals. For $s=200$ (i.e., $40\%$ of the total training budget), we observe that the accuracy of autodiff normals is comparable to the fine-tuned model, which is trained for a total of 800 steps ($160\%$ of the training budget). For $s=0$, i.e., applying \conditionalLink{eq:ftloss}{Eq. (3)} from the first step leads to unstable optimization, causing the angle error to explode. Interestingly, for as low as $s=50$ ($10\%$ of training budget) training steps, we observe that the accuracy of autodiff normals improves compared to the pre-trained model.

This analysis shows that our proposed loss function (\conditionalLink{eq:ftloss}{Eq. (3)}), can also be used to train hybrid neural fields from scratch such that they have more accurate spatial autodiff gradients. This requires an initial warm-start phase where we train the network to fit the zeroth-order signal followed by training with our proposed loss function (\conditionalLink{eq:ftloss}{Eq. (3)}). A higher number of warm-start steps leads to more accurate autodiff normals.

\section{Additional Results for Rendering}
\label{sec:addlrendering}

\begin{figure*}[!h]
    \centering
        \includegraphics[width=0.95\linewidth]{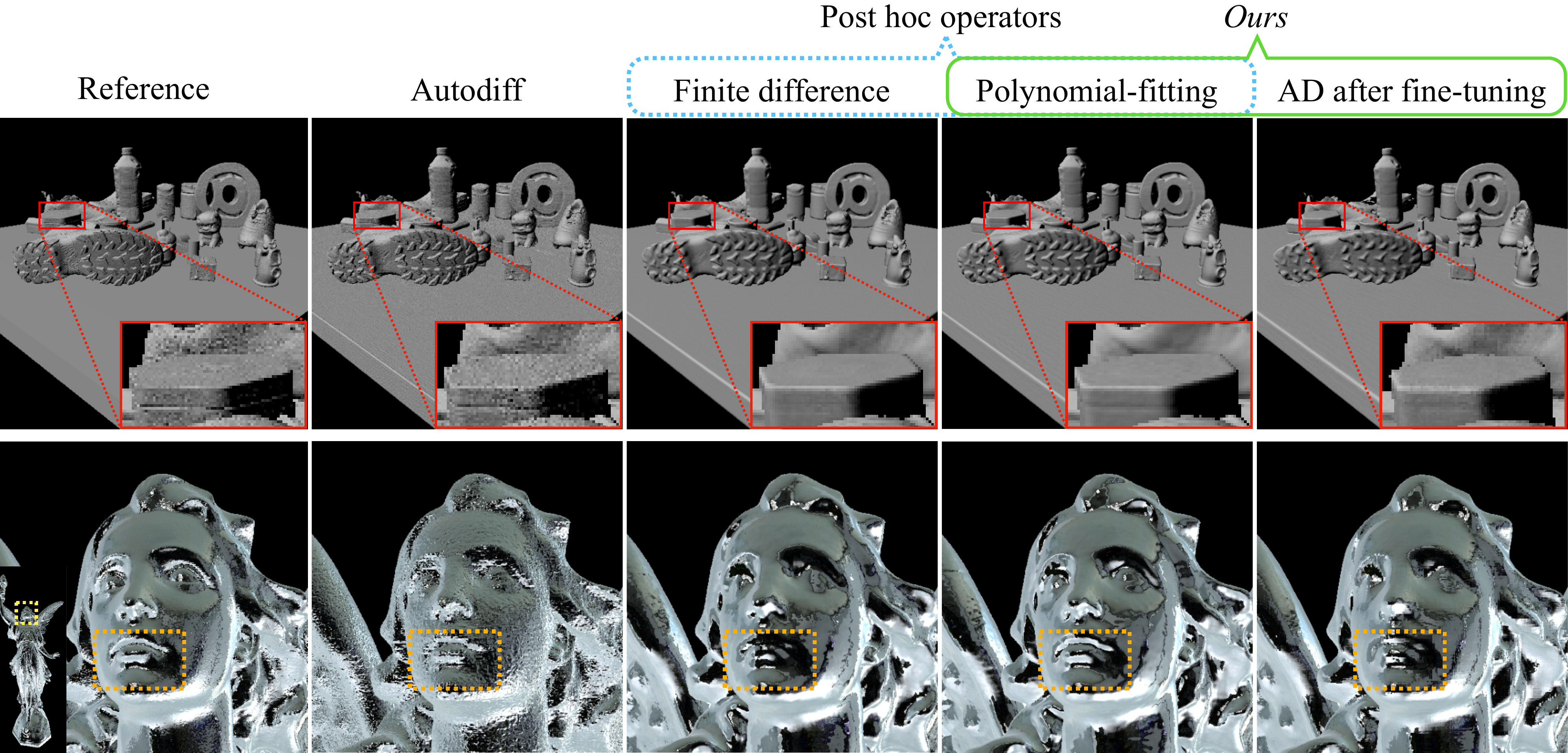}
    \vspace{-0.5em}
    \caption{\textbf{Additional results for rendering.} A large-scale scene lit by a light source put in front of it (top) and specular Lucy  (inset) lit by an environment map (bottom).}
    \label{fig:renderingsuppl}
\end{figure*}

In this section, we provide some additional results for rendering. \Cref{fig:renderingsuppl} shows our results on a large-scale scene (top) and a complex shape (bottom). We observe that our post hoc operator is relatively better at preserving sharp details, such as the boundary between the lid and the box (top), and the contours of the lips (bottom) while reducing the noisy artifacts caused by autodiff surface normals. Our fine-tuning approach also improves the accuracy of autodiff surface normals.

\begin{table*}[h!]
    \centering
    \renewcommand{\arraystretch}{1.2}
    \renewcommand{\tabcolsep}{1.2mm}
    \resizebox{\linewidth}{!}{\begin{tabular}{lcccccccccccccccccc}
\toprule                                      

                \multirow{2}{*}{Shape} & \multicolumn{14}{c}{Surface Normals} & \multicolumn{4}{c}{Mean Curvature} \\
                \cmidrule(lr){2-15} \cmidrule(lr){16-19}
                & \multicolumn{3}{c}{L2 $\downarrow$} & \multicolumn{3}{c}{Ang $\downarrow$} & \multicolumn{3}{c}{AA@1 $\uparrow$} & \multicolumn{3}{c}{AA@2 $\uparrow$} & $\sigma$ &     $h$ & \multicolumn{2}{c}{RRE $\downarrow$} & $\sigma$ &     $h$ \\
                \cmidrule(lr){2-4} \cmidrule(lr){5-7} \cmidrule(lr){8-10} \cmidrule(lr){11-13} \cmidrule(lr){16-17}
                    &   AD &   FD & Ours &    AD &    FD &  Ours &   AD &    FD &  Ours &   AD &    FD & \multicolumn{3}{l}{Ours} &    FD & \multicolumn{3}{l}{Ours} \\
\midrule
        Angel & 0.12 & 0.04 & \textbf{0.03} &  7.14 &  2.30 &  \textbf{1.50} & 1.93 & 26.47 & \textbf{52.09} & 7.50 & 61.66 & \textbf{81.56} &     1.5e-3 & $2 / 512$ &  1.60 & \textbf{1.47} &     9.5e-1 & $2 / 256$ \\
          Armadillo & 0.13 & 0.03 & \textbf{0.02} &  7.27 &  1.83 & \textbf{ 1.18} & 1.79 & 24.69 & \textbf{52.52} & 6.88 & 63.93 & \textbf{86.84} &     2.0e-3 & $2 / 512$ &  1.66 & \textbf{0.81} &     1.5e-1 & $2 / 512$ \\
              
              Bunny & 0.12 & 0.03 & \textbf{0.02} &  6.95 &  1.79 &  \textbf{1.26} & 2.13 & 42.46 &\textbf{ 67.98} & 8.19 & 78.31 & \textbf{88.39} &     5.0e-3 & $2 / 256$ &  1.37 & \textbf{0.72} &     2.5e-1 & $2 / 256$ \\
             Column & 0.72 & 0.28 & \textbf{0.15} & 46.15 & 16.27 &  \textbf{8.47} & 0.31 &  0.51 &  \textbf{4.54} & 1.20 &  2.08 & \textbf{15.87} &     3.5e-3 & $2 / 256$ &  2.54 & \textbf{0.88} &     4.5e-1 & $2 / 128$ \\
             Cup & 0.12 & 0.02 &\textbf{ 0.01} &  7.06 &  1.24 & \textbf{ 0.88} & 2.02 & 62.26 & \textbf{72.20} & 7.93 & 84.37 & \textbf{88.66} &     8.0e-3 & $2 / 128$ &  4.59 & \textbf{0.83} &     2.0e-2 &  $2 / 64$ \\
             Dragon & 0.11 & 0.03 & \textbf{0.02} &  6.45 &  1.88 &  \textbf{1.36} & 2.32 & 29.22 &\textbf{ 54.68} & 8.86 & 69.00 & \textbf{86.67} &     2.0e-3 & $2 / 512$ &  1.46 & \textbf{0.89} &     9.0e-1 & $2 / 256$ \\
             Flower & 0.26 & 0.08 & \textbf{0.06} & 15.21 &  4.50 &  \textbf{3.39} & 0.63 & 39.12 & \textbf{57.18} & 2.52 & 66.99 & \textbf{69.30} &     1.0e-2 & $2 / 128$ & 13.40 & \textbf{0.87} &     2.0e-2 & $2 / 512$ \\
             Galera & 0.12 & 0.04 & \textbf{0.03} &  7.10 &  2.10 &  \textbf{1.65} & 1.82 & 21.89 & \textbf{37.75} & 7.00 & 58.76 & \textbf{75.89} &     2.0e-3 & $2 / 512$ &  1.85 & \textbf{0.82} &     2.5e-1 & $2 / 512$ \\
               Hand & 0.14 & 0.04 & \textbf{0.02} &  8.03 &  2.12 &  \textbf{1.44} & 1.40 & 19.60 & \textbf{39.32} & 5.54 & 55.55 & \textbf{79.82} &     1.5e-3 & $2 / 512$ &  1.27 & \textbf{0.86} &     3.5e-1 & $2 / 256$ \\
            Netsuke & 0.12 & 0.04 & \textbf{0.03} &  7.00 &  2.19 &  \textbf{1.67} & 1.89 & 21.48 & \textbf{41.84} & 7.21 & 56.91 & \textbf{74.48} &     2.0e-3 & $2 / 512$ &  2.40 & \textbf{0.82} &     3.5e-1 & $2 / 512$ \\
            Serapis & 0.11 & \textbf{0.03 }& \textbf{0.03} &  6.59 &  1.88 &  \textbf{1.53} & 2.24 & 36.02 & \textbf{49.25} & 8.60 & 68.78 & \textbf{76.18} &   4.0e-3 & $2 / 256$ & 1.91 & \textbf{0.92} &     2.5e-2 & $2 / 128$ \\
            Tortuga & 0.11 & 0.03 & \textbf{0.02} &  6.07 &  1.51 &  \textbf{1.08} & 2.51 & 46.14 & \textbf{63.30} & 9.70 & 80.90 & \textbf{89.30} &     3.0e-3 & $2 / 256$ &  2.36 & \textbf{0.74} &     2.0e-1 & $2 / 512$ \\
            Utah Teapot & 0.14 & 0.04 & \textbf{0.03} &  8.33 &  2.53 & \textbf{ 2.00} & 1.51 & 27.91 & \textbf{42.20} & 5.91 & 62.64 & \textbf{73.89} &     4.5e-3 & $2 / 256$ &  7.68 & \textbf{0.76} &     3.5e-2 & $2 / 512$ \\
         XYZ Dragon & 0.16 & 0.11 & \textbf{0.07} &  9.19 &  6.37 &  \textbf{4.11} & 1.09 &  4.40 & \textbf{ 6.83} & 4.37 & 15.56 & \textbf{23.86} &     8.0e-4 & $2 / 512$ &  2.00 & \textbf{0.92} &     1.5e-1 & $2 / 512$ \\
      XYZ Statuette & 0.61 & 0.25 & \textbf{0.18} & 37.50 & 14.53 & \textbf{10.55} & 0.12 &  0.72 &  \textbf{2.11} & 0.46 &  2.94 &  \textbf{7.86} &     1.5e-3 & $2 / 512$ &  9.00 & \textbf{0.97} &     2.0e-3 & $2 / 512$ \\

\midrule
               Mean & 0.21 & 0.07 & \textbf{0.05} & 12.40 &  4.20 & \textbf{ 2.80} & 1.58 & 26.86 & \textbf{42.92} & 6.12 & 55.22 & \textbf{67.90} &        - &       - & 3.67 & \textbf{0.89} &        - &       - \\
\bottomrule
\end{tabular}}
\caption{\textbf{Post hoc operator evluation for Instant NGP}. We compare our operators on the FamousShape dataset \cite{ErlerEtAl:Points2Surf:ECCV:2020}. $\sigma, h$ indicate the selected hyperparameters for our approach and finite difference (FD) respectively. Note that our approach provides more accurate derivatives than the baselines.}
\label{tab:opaccfull}
\end{table*}

\begin{table*}[h!]
\small
    \centering
    \renewcommand{\arraystretch}{1.2}
    \renewcommand{\tabcolsep}{1.2mm}
    \resizebox{\linewidth}{!}{
    \begin{tabular}{lccccccccccccccc}
\toprule
               Shape & \multicolumn{6}{c}{Before fine-tuning} & \multicolumn{6}{c}{After fine-tuning} & $\sigma$ \\
               \cmidrule(lr){2-7} \cmidrule(lr){8-13}
                     &   L2 $\downarrow$ & Ang $\downarrow$ & AA@1 $\uparrow$ & AA@2 $\uparrow$ &  CD $\downarrow$ & F-Score $\uparrow$ & L2 $\downarrow$ &  Ang $\downarrow$ &  AA@1 $\uparrow$ & AA@2 $\uparrow$ &       CD $\downarrow$ & F-Score $\uparrow$ &  \\
\midrule
               Angel &      0.12 &    7.13 & 1.92 & 7.54 &   5.32e-4 &       93.47 &     \textbf{0.04} &    \textbf{2.06} & \textbf{31.74} & \textbf{67.25} &  5.38e-4 &       93.45 &   1.5e-3 \\
           Armadillo &      0.13 &    7.23 & 1.80 & 6.95 & 1.63e-4 &       96.15 &     \textbf{0.03} &    \textbf{1.72} & \textbf{31.04} & \textbf{69.70} &  1.65e-4 &       96.14 &   2.0e-3 \\
               Bunny &      0.12 &    6.98 & 2.08 & 8.11 & 7.26e-4 &       93.26 &     \textbf{0.02} &    \textbf{1.37} & \textbf{60.74} & \textbf{86.52} &  7.09e-4 &       93.35 &   5.5e-3 \\
              Column &      0.73 &   46.25 & 0.31 & 1.24 &  2.93e-3 &       85.89 &     \textbf{0.14} &    \textbf{8.35} &  \textbf{4.66} & \textbf{16.16} & 2.95e-3 &       85.71 &   3.5e-3 \\
                 Cup &      0.12 &    7.05 & 2.06 & 7.91 &  3.24e-4 &       94.47 &     \textbf{0.02} &    \textbf{1.15} & \textbf{60.76} & \textbf{84.76} &  3.27e-4 &       89.11 &   1.0e-2 \\
              Dragon &      0.11 &    6.46 & 2.31 & 8.81 & 1.99e-3 &       89.97 &     \textbf{0.03} &    \textbf{1.90} & \textbf{33.25} & \textbf{70.87} & 1.98e-3 &       89.96 &   2.0e-3 \\
              Flower &      0.26 &   15.20 & 0.67 & 2.52 &  3.40e-4 &       96.48 &     \textbf{0.06} &    \textbf{3.32} & \textbf{55.34} & \textbf{70.17} & 3.47e-4 &       91.49 &   1.0e-2 \\
              Galera &      0.12 &    7.10 & 1.82 & 6.93 &  8.37e-4 &       92.29 &     \textbf{0.03} &    \textbf{1.97} & \textbf{28.11} & \textbf{65.63} & 8.35e-4 &       92.31 &   2.0e-3 \\
                Hand &      0.14 &    8.02 & 1.39 & 5.53 &  2.64e-3 &       88.08 &     \textbf{0.04} &    \textbf{2.10} & \textbf{21.30} & \textbf{57.41} &  2.67e-3 &       88.10 &   1.5e-3 \\
             Netsuke &      0.12 &    7.01 & 1.90 & 7.29 & 1.86e-4 &       96.13 &     \textbf{0.04} &    \textbf{2.11} & \textbf{26.40} & \textbf{61.64} &   1.87e-4 &       96.11 &   2.0e-3 \\
             Serapis &      0.11 &    6.57 & 2.21 & 8.55 &   1.18e-3 &       91.79 &     \textbf{0.03} &    \textbf{1.65} & \textbf{44.96} & \textbf{73.21} & 1.18e-3 &       91.73 &   4.0e-3 \\
             Tortuga &      0.11 &    6.04 & 2.53 & 9.75 &  3.29e-4 &       96.04 &     \textbf{0.02} &    \textbf{1.18} & \textbf{61.18} & \textbf{87.00} &  3.28e-4 &       96.07 &   3.5e-3 \\
 Utah Teapot &      0.14 &    8.29 & 1.50 & 5.92 & 6.23e-4 &       94.30 &     \textbf{0.04} &    \textbf{2.06} & \textbf{38.10} & \textbf{70.07} &  6.31e-4 &       94.13 &   4.0e-3 \\
          XYZ Dragon &      0.16 &    9.18 & 1.13 & 4.40 &  9.72e-4 &       90.40 &     \textbf{0.10} &    \textbf{5.81} &  \textbf{4.62} & \textbf{16.47} & 9.72e-4 &       90.37 &   1.5e-3 \\
       XYZ Statuette &      0.61 &   37.46 & 0.13 & 0.46 &  9.69e-5 &       97.29 &     \textbf{0.19} &   \textbf{11.11} &  \textbf{1.77} &  \textbf{6.76} &   9.97e-5 &       96.17 &   1.5e-3 \\
\midrule
        Mean & 0.21 & 12.38 & 1.58 & 6.12 & 9.24e-4 & 93.07 & \textbf{0.05} & \textbf{3.20} & \textbf{33.59} & \textbf{60.24} & 9.28e-4 & 92.28 & - \\
\bottomrule
\end{tabular}}
\caption{\textbf{Fine-tuning using polynomial-fitting for Instant NGP}. Full results for fine-tuning using polynomial-fitting over the FamousShape dataset \cite{ErlerEtAl:Points2Surf:ECCV:2020}. $\sigma$ denotes the hyperparameter value with the best results from the ensemble.}
    \label{tab:ftresfull}
\end{table*}

\begin{table*}[h!]
    \centering
    \renewcommand{\arraystretch}{1.2}
    \renewcommand{\tabcolsep}{1.2mm}
    \resizebox{\linewidth}{!}{{\begin{tabular}{lccccccccccccccccccc}
\toprule                                      

                \multirow{2}{*}{Shape} & \multicolumn{14}{c}{Surface Normals} & \multicolumn{5}{c}{Mean Curvature} \\
                \cmidrule(lr){2-15} \cmidrule(lr){16-20}
                & \multicolumn{3}{c}{L2 $\downarrow$} & \multicolumn{3}{c}{Ang $\downarrow$} & \multicolumn{3}{c}{AA@1 $\uparrow$} & \multicolumn{3}{c}{AA@2 $\uparrow$} & $\sigma$ &     $h$ & \multicolumn{3}{c}{RRE $\downarrow$} & $\sigma$ &     $h$ \\
                \cmidrule(lr){2-4} \cmidrule(lr){5-7} \cmidrule(lr){8-10} \cmidrule(lr){11-13} \cmidrule(lr){16-18}
                    &   AD &   FD & Ours &    AD &    FD &  Ours &   AD &    FD &  Ours &   AD &    FD & \multicolumn{3}{l}{Ours} &              \multicolumn{2}{c}{FD}& \multicolumn{3}{l}{Ours} \\
\midrule
        Angel & 0.09 & 0.05 & \textbf{0.04} &  5.20 &  2.71 &  \textbf{2.39} & 13.56 & 33.30 & \textbf{43.95} & 33.37 & 58.60 & \textbf{66.99 }&     2.0e-3 & $2 / 512$ & \multicolumn{2}{c}{3.41}& \textbf{0.87} &     4.5e-1 & $2 / 256$ \\
          Armadillo & 0.08 & 0.04 & \textbf{0.03} &  4.87 &  2.09 &  \textbf{1.75} &  7.84 & 24.00 & \textbf{32.72} & 23.34 & 58.00 & \textbf{68.52} &     2.0e-3 & $2 / 512$ & \multicolumn{2}{c}{1.75}& \textbf{1.49} &     9.0e-1 & $2 / 256$ \\
              Bunny & 0.07 & 0.03 & \textbf{0.02} &  3.78 &  1.73 &  \textbf{1.26} & 13.10 & 47.91 & \textbf{67.93} & 37.00 & 79.02 & \textbf{88.29} &     5.0e-3 & $2 / 256$ & \multicolumn{2}{c}{1.25}& \textbf{0.81} &     1.5e-1 & $2 / 256$ \\
             Column & 0.27 & 0.21 & \textbf{0.14} & 16.07 & 11.98 &  \textbf{8.33} &  1.96 &  4.09 & \textbf{11.48} &  6.96 & 12.64 & \textbf{24.65} &     3.5e-3 & $2 / 512$ & \multicolumn{2}{c}{4.62}& \textbf{0.83} &     2.0e-2 &  $2 / 64$ \\
                Cup & 0.06 & 0.02 & \textbf{0.01} &  3.45 &  1.20 &  \textbf{0.86} & 21.27 & 64.09 & \textbf{72.44} & 45.64 & 84.11 & \textbf{88.60} &     8.0e-3 & $2 / 128$ & \multicolumn{2}{c}{1.24}& \textbf{0.72} &     2.5e-1 & $2 / 256$ \\
             Dragon & 0.08 & 0.04 & \textbf{0.03} &  4.56 &  2.25 &  \textbf{1.76} & 10.73 & 26.83 & \textbf{44.20} & 30.92 & 60.64 & \textbf{76.86} &     2.5e-3 & $2 / 512$ & \multicolumn{2}{c}{1.52}& \textbf{0.89} &     9.0e-1 & $2 / 256$ \\
             Flower & 0.14 & 0.07 & \textbf{0.06} &  8.13 &  4.26 &  \textbf{3.36} & 14.50 & \textbf{57.72} & 57.60 & 37.26 & \textbf{70.87} & 69.32 &     1.0e-2 & $2 / 128$ & \multicolumn{2}{c}{3.28}& \textbf{0.87} &     2.0e-2 &  $2 / 32$ \\
             Galera & 0.08 & 0.04 & \textbf{0.04} &  4.62 &  2.40 &  \textbf{2.07} & 10.22 & 23.97 & \textbf{32.01} & 29.11 & 55.73 & \textbf{65.33} &     2.0e-3 & $2 / 512$ & \multicolumn{2}{c}{1.70}& \textbf{0.82} &     2.5e-1 & $2 / 512$ \\
               Hand & 0.09 & 0.04 & \textbf{0.04} &  4.93 &  2.55 &  \textbf{2.03} &  8.24 & 19.34 & \textbf{26.81} & 25.71 & 50.25 & \textbf{62.64} &     2.0e-3 & $2 / 512$ & \multicolumn{2}{c}{1.90}& \textbf{0.86} &     3.5e-1 &  $2 / 64$ \\
            Netsuke & 0.08 & 0.04 & \textbf{0.03} &  4.56 &  2.26 &  \textbf{1.99} & 10.12 & 26.22 & \textbf{33.08} & 28.77 & 57.91 & \textbf{65.06} &     2.0e-3 & $2 / 512$ & \multicolumn{2}{c}{1.45}& \textbf{0.82} &     3.5e-1 & $2 / 256$ \\
            Serapis & 0.07 & 0.03 & \textbf{0.03} &  4.01 &  1.89 &  \textbf{1.54} & 17.73 & 39.21 & \textbf{48.89} & 36.91 & 67.65 & \textbf{75.46} &     4.0e-3 & $2 / 256$ & \multicolumn{2}{c}{1.98}& \textbf{0.93} &     2.5e-2 & $2 / 128$ \\
            Tortuga & 0.05 & 0.03 & \textbf{0.02} &  2.94 &  1.47 &  \textbf{1.13} & 17.44 & 50.36 & \textbf{62.65} & 45.51 & 80.91 & \textbf{87.94} &     3.0e-3 & $2 / 256$ & \multicolumn{2}{c}{1.60}& \textbf{0.74} &     2.0e-1 & $2 / 512$ \\
            Utah Teapot & 0.06 & 0.04 & \textbf{0.03} &  3.51 &  2.28 &  \textbf{1.98} & 22.17 & 36.37 & \textbf{42.62} & 47.55 & 67.44 & \textbf{73.79} &     4.5e-3 & $2 / 256$ & \multicolumn{2}{c}{0.96}& \textbf{0.76} &     3.5e-2 &  $2 / 32$ \\
         XYZ Dragon & 0.16 & 0.13 & \textbf{0.12} &  9.39 &  7.37 &  \textbf{6.89} &  2.16 &  3.58 &  \textbf{4.00} &  8.15 & 12.72 & \textbf{14.01} &     1.5e-3 & $2 / 512$ & \multicolumn{2}{c}{2.13}& \textbf{0.92} &     1.5e-1 & $2 / 256$ \\
      XYZ Statuette & 0.31 & 0.23 & \textbf{0.21} & 18.26 & 13.13 & \textbf{12.27} &  1.36 &  2.91 &  \textbf{3.88} &  4.82 &  9.41 & \textbf{12.28} &     1.5e-3 & $2 / 512$ & \multicolumn{2}{c}{10.54}& \textbf{0.97} &     2.5e-3 & $2 / 512$ \\
      \midrule
      Mean & 0.11 & 0.07 & \textbf{0.06} &  6.55 &  3.97 &  \textbf{3.31} & 11.49 & 30.66 & \textbf{38.95} & 29.40 & 55.06 & \textbf{62.65} &        - &       - & \multicolumn{2}{c}{2.62}& \textbf{0.89} &        - &       - \\
\bottomrule
\end{tabular}}}
\caption{{\textbf{Post hoc operator evluation on Dense Grid}. Comparison on the FamousShape dataset \cite{ErlerEtAl:Points2Surf:ECCV:2020}. $\sigma, h$ indicate the selected hyperparameters for our approach and finite difference (FD) respectively. Note that our approach provides more accurate surface normals and mean curvature than the baselines.}}
    \label{tab:opaccdense}
\end{table*}

\begin{table*}[h!]
\small
    \centering
    \renewcommand{\arraystretch}{1.2}
    \renewcommand{\tabcolsep}{1.2mm}
    \resizebox{\linewidth}{!}{{
    \begin{tabular}{lccccccccccccc}
\toprule
               Shape & \multicolumn{6}{c}{Before fine-tuning} & \multicolumn{6}{c}{After fine-tuning} & $\sigma$ \\
               \cmidrule(lr){2-7} \cmidrule(lr){8-13}
                     &   L2 $\downarrow$ & Ang $\downarrow$ & AA@1 $\uparrow$ & AA@2 $\uparrow$ &   CD $\downarrow$ & F-Score $\uparrow$ & L2 $\downarrow$ &  Ang $\downarrow$ &  AA@1 $\uparrow$ & AA@2 $\uparrow$ &       CD $\downarrow$ & F-Score $\uparrow$ &  \\
\midrule
                             Angel &      0.09 &    5.20 & 13.64 & 33.45 &        5.33e-4 &      92.87 &     \textbf{0.06} &    \textbf{3.62} & \textbf{30.12} & \textbf{53.67} & 5.35e-4 &      92.50 & 2.0e-3\\
          Armadillo &      0.08 &    4.87 &  7.72 & 23.40 & 1.65e-4 &      95.28 &     \textbf{0.06} &    \textbf{3.27} & \textbf{14.84} & \textbf{39.06} &  1.69e-4 &      95.02 & 2.0e-3\\
              Bunny &      0.07 &    3.78 & 13.11 & 37.07 &  7.25e-4 &      91.00 &     \textbf{0.03} &    \textbf{1.59} & \textbf{52.63} & \textbf{81.11} & 7.15e-4 &      90.67 & 5.0e-3\\
             Column &      0.27 &   16.15 &  1.87 &  6.90 & 2.95e-3 &      84.82 &     \textbf{0.17} &    \textbf{9.79} &  \textbf{4.43} & \textbf{13.75} & 2.92e-3 &      73.45 & 3.5e-3\\
                Cup &      0.06 &    3.48 & 21.14 & 45.34 & 3.24e-4 &      84.89 &     \textbf{0.02} &    \textbf{1.11} & \textbf{63.64} & \textbf{84.39} & 3.20e-4 &      78.32 & 8.0e-3\\
             Dragon &      0.08 &    4.54 & 10.74 & 30.85 & 1.99e-3 &      86.31 &     \textbf{0.05} &    \textbf{2.72} & \textbf{26.48} & \textbf{57.89} & 1.99e-3 &      85.95 & 2.5e-3\\
             Flower &      0.14 &    8.14 & 14.34 & 37.25 & 3.40e-4 &      91.50 &     \textbf{0.06} &    \textbf{3.40} & \textbf{54.03} & \textbf{68.48} & 3.46e-4 &      83.98 & 1.0e-2\\
             Galera &      0.08 &    4.62 & 10.08 & 28.87 & 8.41e-4 &      85.93 &     \textbf{0.05} &    \textbf{2.93} & \textbf{23.18} & \textbf{51.34} & 8.36e-4 &      85.87 & 2.0e-3\\
               Hand &      0.09 &    4.95 &  8.16 & 25.72 & 2.64e-3 &      87.68 &     \textbf{0.06} &    \textbf{3.28} & \textbf{13.77} & \textbf{39.75} & 2.65e-3 &      87.59 & 2.0e-3\\
            Netsuke &      0.08 &    4.55 & 10.21 & 29.01 & 1.86e-4 &      92.93 &     \textbf{0.05} &    \textbf{2.93} & \textbf{20.56} & \textbf{47.99} & 1.84e-4 &      92.82 & 2.0e-3\\
            Serapis &      0.07 &    4.01 & 17.48 & 36.62 & 1.18e-3 &      85.15 &    \textbf{ 0.03} &    \textbf{1.97} & \textbf{41.59} & \textbf{66.92} & 1.17e-3 &      84.85 & 4.0e-3\\
            Tortuga &      0.05 &    2.94 & 17.30 & 45.35 & 3.29e-4 &      93.16 &     \textbf{0.03} &    \textbf{1.47} & \textbf{51.17} & \textbf{80.15} & 3.30e-4 &      93.04 & 3.0e-3\\
Utah Teapot &      0.06 &    3.52 & 22.07 & 47.55 & 6.22e-4 &      90.70 &     \textbf{0.04} &    \textbf{2.19} & \textbf{39.15} & \textbf{71.07} & 6.24e-4 &      89.93 & 4.5e-3\\
         XYZ Dragon &      0.16 &    9.38 &  2.17 &  8.11 & 9.72e-4 &      89.68 &     \textbf{0.15} &    \textbf{8.66} &  \textbf{2.82} & \textbf{10.23} & 9.77e-4 &      89.28 & 1.5e-3\\
      XYZ Statuette &      0.31 &   18.23 &  1.34 &  4.89 & 9.70e-5 &      95.58 &    \textbf{ 0.26} &   \textbf{15.33} &  \textbf{2.20} &  \textbf{7.55} & 1.03e-4 &      91.53 & 1.5e-3\\
      \midrule
                Mean &      0.11 &    6.56 & 11.42 & 29.35 &   9.26e-4 &      89.83 &     \textbf{0.08} &    \textbf{4.40} & \textbf{29.32} & \textbf{51.40} &     9.25e-4 &      87.66 & - \\ 
\bottomrule
\end{tabular}}}
\caption{\textbf{Fine-tuning using polynomial-fitting on Dense Grid}. Full results for fine-tuning using polynomial-fitting over the FamousShape dataset \cite{ErlerEtAl:Points2Surf:ECCV:2020}. $\sigma$ denotes the hyperparameter value that obtained the best results.}
    \label{tab:ftresdense}
\end{table*}

\begin{table*}[h!]
    \centering
    \renewcommand{\arraystretch}{1.2}
    \renewcommand{\tabcolsep}{1.2mm}
    \resizebox{\linewidth}{!}{\begin{tabular}{lcccccccccccccccccc}
\toprule                                      

                \multirow{2}{*}{Shape} & \multicolumn{14}{c}{Surface Normals} & \multicolumn{4}{c}{Mean Curvature} \\
                \cmidrule(lr){2-15} \cmidrule(lr){16-19}
                & \multicolumn{3}{c}{L2 $\downarrow$} & \multicolumn{3}{c}{Ang $\downarrow$} & \multicolumn{3}{c}{AA@1 $\uparrow$} & \multicolumn{3}{c}{AA@2 $\uparrow$} & $\sigma$ &     $h$ & \multicolumn{2}{c}{RRE $\downarrow$} & $\sigma$ &     $h$ \\
                \cmidrule(lr){2-4} \cmidrule(lr){5-7} \cmidrule(lr){8-10} \cmidrule(lr){11-13} \cmidrule(lr){16-17}
                    &   AD &   FD & Ours &    AD &    FD &  Ours &   AD &    FD &  Ours &   AD &    FD & \multicolumn{3}{l}{Ours} &     FD & \multicolumn{3}{l}{Ours} \\
\midrule
        Angel & 0.08 & 0.04 & \textbf{0.03} &  4.86 &  2.30 &  \textbf{1.92} & 5.91 & 27.50 & \textbf{30.46} & 20.85 & 62.71 & \textbf{69.31} &     1.5e-3 & $2 / 512$ &  2.99 & \textbf{1.63} &     9.0e-1 & $2 / 512$ \\
          Armadillo & 0.09 & 0.03 & \textbf{0.03} &  5.35 &  2.00 &  \textbf{1.48} & 4.04 & 21.41 & \textbf{35.38} & 15.00 & 58.19 & \textbf{77.42} &     2.0e-3 & $2 / 512$ &  1.23 & \textbf{0.81} &     2.0e-1 & $2 / 256$ \\
              Bunny & 0.10 & 0.03 & \textbf{0.02} &  5.74 &  1.98 &  \textbf{1.31} & 3.95 & 33.16 & \textbf{65.12} & 14.66 & 71.98 & \textbf{87.97} &     5.0e-3 & $2 / 256$ &  1.43 & \textbf{0.72} &     2.5e-1 & $2 / 256$ \\
             Column & 0.38 & 0.24 & \textbf{0.14} & 22.87 & 14.18 &  \textbf{8.31} & 0.70 &  2.06 &  \textbf{8.21} &  2.68 &  7.60 & \textbf{23.31} &     3.0e-3 & $2 / 512$ &  6.07 & \textbf{0.95} &     3.0e-2 & $2 / 512$ \\
                Cup & 0.09 & 0.02 & \textbf{0.02} &  5.20 &  1.31 &  \textbf{0.90} & 4.86 & 57.47 & \textbf{71.78} & 17.38 & 83.76 & \textbf{88.42} &     8.0e-3 & $2 / 128$ &  6.32 & \textbf{0.82} &     9.0e-4 &  $2 / 32$ \\
             Dragon & 0.09 & 0.04 & \textbf{0.03} &  5.24 &  2.32 &  \textbf{1.77} & 4.56 & 19.24 & \textbf{36.44} & 16.54 & 53.50 & \textbf{75.71} &     2.5e-3 & $2 / 512$ &  1.58 & \textbf{0.93} &     9.0e-1 & $2 / 256$ \\
             Flower & 0.18 & 0.08 & \textbf{0.06} & 10.65 &  4.47 &  \textbf{3.40} & 3.08 & 44.50 & \textbf{56.91} & 11.46 & 69.30 & \textbf{69.07} &     1.0e-2 & $2 / 128$ &  2.68 & \textbf{0.87} &     2.0e-2 &  $2 / 64$ \\
             Galera & 0.10 & 0.04 & \textbf{0.03} &  6.01 &  2.51 &  \textbf{1.97} & 3.25 & 15.84 & \textbf{24.95} & 12.33 & 46.93 & \textbf{63.97} &     2.0e-3 & $2 / 512$ &  1.37 & \textbf{0.82} &     2.5e-1 & $2 / 256$ \\
               Hand & 0.08 & 0.04 & \textbf{0.03} &  4.46 &  2.03 &  \textbf{1.59} & 6.42 & 22.34 & \textbf{33.03} & 22.43 & 59.48 & \textbf{74.47} &     1.5e-3 & $2 / 512$ &  1.88 & \textbf{0.85} &     3.5e-1 &  $2 / 64$ \\
               
            Netsuke & 0.10 & 0.04 & \textbf{0.04} &  5.92 &  2.52 &  \textbf{2.05} & 3.47 & 16.26 & \textbf{25.08} & 12.94 & 47.23 & \textbf{62.22} &     2.0e-3 & $2 / 512$ &  1.55 & \textbf{0.82} &     3.5e-1 & $2 / 256$ \\
            Serapis & 0.11 & 0.04 & \textbf{0.03} &  6.24 &  2.17 &  \textbf{1.65} & 3.47 & 25.35 & \textbf{45.18} & 12.82 & 60.02 & \textbf{74.30} &     4.0e-3 & $2 / 256$ &  6.32 & \textbf{0.93} &     2.5e-2 & $2 / 512$ \\
            Tortuga & 0.09 & 0.03 & \textbf{0.02} &  5.32 &  1.76 &  \textbf{1.23} & 4.36 & 34.08 & \textbf{57.78} & 15.83 & 72.24 & \textbf{87.04} &     3.5e-3 & $2 / 256$ &  2.86 & \textbf{0.74} &     2.0e-1 & $2 / 512$ \\
        Utah Teapot & 0.11 & 0.05 & \textbf{0.04} &  6.16 &  2.62 &  \textbf{2.04} & 5.02 & 28.41 & \textbf{40.87} & 17.91 & 61.90 & \textbf{73.30} &     4.5e-3 & $2 / 256$ &  7.03 & \textbf{0.75} &     3.5e-2 & $2 / 512$ \\
         XYZ Dragon & 0.22 & 0.13 & \textbf{0.12} & 12.80 &  7.52 &  \textbf{6.96} & 0.72 &  2.40 &  \textbf{2.43} &  2.87 &  9.12 &  \textbf{9.22} &     1.5e-3 & $2 / 512$ &  2.78 & \textbf{0.92} &     1.5e-1 & $2 / 512$ \\
      XYZ Statuette & 0.37 & 0.23 & \textbf{0.21} & 22.04 & 13.16 & \textbf{11.91} & 0.31 &  1.29 &  \textbf{1.35} &  1.27 &  5.13 &  \textbf{5.29} &     1.5e-3 & $2 / 512$ & 15.64 & \textbf{0.97} &     2.5e-3 & $2 / 512$ \\
      \midrule
      Mean & 0.15 & 0.07 & \textbf{0.06} &  8.59 &  4.19 &  \textbf{3.23} & 3.61 & 23.42 & \textbf{35.67} & 13.13 & 51.27 & \textbf{62.74} &        - &       - &  4.12 & \textbf{0.90} &        - &       - \\
\bottomrule
\end{tabular}}
\caption{{\textbf{Post hoc operator evluation on Tri-planes}. Comparison on the FamousShape dataset \cite{ErlerEtAl:Points2Surf:ECCV:2020}. $\sigma, h$ indicate the selected hyperparameters for our approach and finite difference (FD) respectively.}}
    \label{tab:opacctri}
\end{table*}

\end{document}